\documentclass{article}

\usepackage{microtype}
\usepackage{graphicx}
\usepackage{subcaption}
\usepackage{booktabs} 
\usepackage{dblfloatfix}
\usepackage{multicol}
\usepackage{multirow}
\usepackage{makecell}
\usepackage{microtype}

\newcommand\Tstrut{\rule{0pt}{2.6ex}}
\newcommand\TTstrut{\rule{0pt}{3.8ex}}
\newcommand\Bstrut{\rule[-0.9ex]{0pt}{0pt}}
\renewcommand{\arraystretch}{0.5} 
\usepackage{hyperref}

\usepackage[preprint]{icml2026}

\usepackage{amsmath}
\usepackage{amssymb}
\usepackage{mathtools}
\usepackage{amsthm}

\usepackage[capitalize,noabbrev]{cleveref}

\renewcommand{\paragraph}[1]{\vspace{.1em}\noindent\textbf{#1}}

\theoremstyle{plain}

\theoremstyle{definition}

\theoremstyle{remark}

\usepackage[textsize=tiny]{todonotes}

\icmltitlerunning{Routing the Lottery: Adaptive Subnetworks for Heterogeneous Data}

\begin{document}

\twocolumn[
  \icmltitle{Routing the Lottery: Adaptive Subnetworks for Heterogeneous Data}



  \icmlsetsymbol{equal}{*}

  \begin{icmlauthorlist}
    \icmlauthor{Grzegorz Stefański}{aic}
    \icmlauthor{Alberto Presta}{aic}
    \icmlauthor{Michał Byra}{aic,pan}
  \end{icmlauthorlist}

  \icmlaffiliation{aic}{Samsung AI Center Warsaw, Poland}
  \icmlaffiliation{pan}{Institute of Fundamental Technological Research, Polish Academy of Sciences, Poland}

  \icmlcorrespondingauthor{Grzegorz Stefański}{g.stefanski@samsung.com}

  \icmlkeywords{Adaptive Pruning, Lottery Ticket Hypothesis, Heterogeneous Data, Context-Aware Inference}

  \vskip 0.3in
]



\printAffiliationsAndNotice{}  

\begin{abstract}
In pruning, the Lottery Ticket Hypothesis posits that large networks contain sparse subnetworks, or winning tickets, that can be trained in isolation to match the performance of their dense counterparts. However, most existing approaches assume a single universal winning ticket shared across all inputs, ignoring the inherent heterogeneity of real-world data. In this work, we propose Routing the Lottery (RTL), an adaptive pruning framework that discovers multiple specialized subnetworks, called adaptive tickets, each tailored to a class, semantic cluster, or environmental condition. Across diverse datasets and tasks, RTL consistently outperforms single- and multi-model baselines in balanced accuracy and recall, while using up to 10 times fewer parameters than independent models and exhibiting semantically aligned. Furthermore, we identify subnetwork collapse, a performance drop under aggressive pruning, and introduce a subnetwork similarity score that enables label-free diagnosis of oversparsification. Overall, our results recast pruning as a mechanism for aligning model structure with data heterogeneity, paving the way toward more modular and context-aware deep learning.
\end{abstract}    
\section{Introduction}
\label{sec:intro}

Despite the remarkable progress deep neural networks have achieved over the past decades, modern models often require billions of parameters and hundreds of gigaflops per inference, making them impractical for deployment in resource-constrained or real-time settings.

This inefficiency stands in stark contrast to biological intelligence, which achieves high performance with extreme parsimony. 
While specialized hardware continues to advance, the growth in model scale consistently outpaces gains in computational efficiency. Consequently, reducing model complexity is not only crucial for practical deployment but also for understanding the fundamental principles of generalization in deep learning.

Pruning has been a cornerstone of model efficiency research \cite{cheng2024survey}. By removing redundant weights or structures, it aims to uncover smaller subnetworks that retain the performance of their dense counterparts. The Lottery Ticket Hypothesis (LTH) \cite{Frankle2018LTH} revitalized this line of work by positing that ``winning tickets'', i.e. sparse subnetworks within large, randomly initialized networks, can be trained in isolation to match full-model accuracy. This reframed pruning as a discovery process aiming to reveal the minimal structures responsible for learning.

However, nearly all LTH-inspired methods assume a universal subnetwork - a single sparse mask applied uniformly across all inputs, overlooking real-world data heterogeneity.
Different classes, clusters, or environmental conditions often rely on distinct feature representations. A one-size-fits-all mask may therefore sacrifice performance by forcing diverse patterns through a shared, rigid architecture. Bridging this gap requires moving beyond global sparsity toward adaptive, data-aware pruning, a shift that aligns model structure with the intrinsic organization of the data itself.

In this work, we introduce \textit{Routing the Lottery (RTL)}, a novel adaptive pruning framework that rethinks LTH by discovering multiple specialized subnetworks, dubbed \textit{adaptive tickets}, each tailored to a distinct data subset (e.g., a class or semantic cluster). This shift enables the model to allocate representational capacity heterogeneously across the input space, aligning sparsity with data structure rather than enforcing uniform compression.
Our framework achieves specialization through pruning alone, without auxiliary routing networks or additional parameters.
It discovers multiple stable, sparse subnetworks, each tied to a class or cluster, and selects them via simple context-based routing (e.g., label or environment), offering a lightweight and interpretable alternative to Mixture-of-Experts (MoE) architectures that prioritizes structural efficiency over dynamic routing flexibility.

This work serves as a precursor, aiming to guide pruning from a static tool into a dynamic mechanism for building modular and semantically grounded models. The major contributions of this paper are:
\begin{itemize}
    \item RTL jointly learns multiple sparse subnetworks from a shared dense initialization, with masks adapted to data subsets while preserving parameter sharing, maintaining a single compact backbone and employing a mask-based routing to enable context-aware inference.
    \item We show that class-specific subnetworks on CIFAR-10 outperform both single-mask and multi-model pruning baselines while using up to an order of magnitude fewer parameters than independent models, and that on CIFAR-100, RTL scales naturally to enable effective specialization across a larger number of classes.
    \item We validate RTL on a real-world speech enhancement task, where subnetworks specialized for acoustic environments achieve higher SI-SNRi than universal or independent baselines. 
\end{itemize}

\section{Related Work}

\paragraph{Pruning} has a rich history, beginning with sensitivity-based methods like Optimal Brain Damage \cite{LeCun1989} and Optimal Brain Surgeon \cite{Hassibi1993}. 
In particular, unstructured pruning removes non-relevant weights without imposing structural constraints, i.e., without removing entire layers. \cite{han2015learning} performed pruning by learning meaningful connections, while \cite{yang2017designing, yang2018netadapt} exploit energy consumption as a metric for removing elements.
\cite{sreenivasan2022rare, tartaglione2022loss} introduced regularization terms to constrain the magnitude of non-relevant parameters during iterative training, while \cite{benbaki2023fast} considered the combined effect of pruning and updating multiple weights under a sparsity constraint. 
Furthermore, \cite{tartaglione2020pruning} compared and analyze two different pruning approaches, i.e. one-shot and gradual, highlighting that the latter allows for better generalization.
\cite{zhang2024sparse} analyzed fundamental aspects of pruning, identifying two key factors that determine the pruning ratio limit, i.e., weight magnitude and network sharpness, while \cite{liao2023can, hur2019entropy, luo2017entropy, min20182pfpce} proposed entropy-based approaches to guide pruning. 
In general, owing to the conceptual simplicity of pruning, a wide range of methods have been proposed for different scenarios, such as LLM pruning \cite{frantar2023sparsegpt, sun2023simple, lu2024spp, wei2024structured, tan2024wrp}, convolutional pruning \cite{zhao2023automatic}, and even spiking neural networks \cite{shi2024towards}. However, none of these have considered whether finding a single mask applicable to all types of data might be a suboptimal solution.

Iterative Magnitude Pruning (IMP) \cite{Frankle2018LTH} emerged as a practical algorithm for identifying winning tickets, though it remains computationally intensive due to repeated training cycles.
Furthermore, \cite{paul2022unmasking} attempted to demystify the IMP method, investigating what kind of information the obtained mask encodes and how SGD allows the network to extract such information. 
Despite numerous extensions covering initialization schemes \cite{Frankle2021}, learning rate rewinding \cite{Renda2020}, theoretical analysis \cite{tartaglione2022rise, burkholz2021existence, sakamoto2022analyzing, paul2022unmasking},  and algorithmic variants \cite{wang2023loft, lin2023memory}, the LTH paradigm has largely focused on global masks shared across the entire dataset.

\paragraph{Dynamic sparse training} methods relax the fixed-mask assumption by evolving sparsity patterns during training. Approaches such as SET \cite{Mocanu2017}, SNFS \cite{Dettmers2020}, and RigL \cite{Evci2020} prune and regrow connections online, implicitly acknowledging that different inputs may activate different network pathways. 
Furthermore, \cite{molchanov2017variational} extended variational dropout in order to sparsify deep neural networks, while \cite{tartaglione2021serene} introduced sensitivity-based regularization of neurons to learn structured sparse topologies, exploiting neural sensitivity as a regularizer. 
Nevertheless, these methods still maintain a single evolving subnetwork rather than explicitly specializing distinct structures for different data regimes.

Closer in spirit are conditional computation techniques like MoE \cite{Shazeer2017, Fedus2022} and conditional convolutions \cite{Yang2019}, which activate input-dependent subnetworks to improve scalability. However, these methods typically require complex routing mechanisms, large auxiliary parameter sets, and substantial compute budgets.

\section{Method}
\label{sec:method}

\begin{figure}[t!]
    \centering
    \includegraphics[width=1\columnwidth]{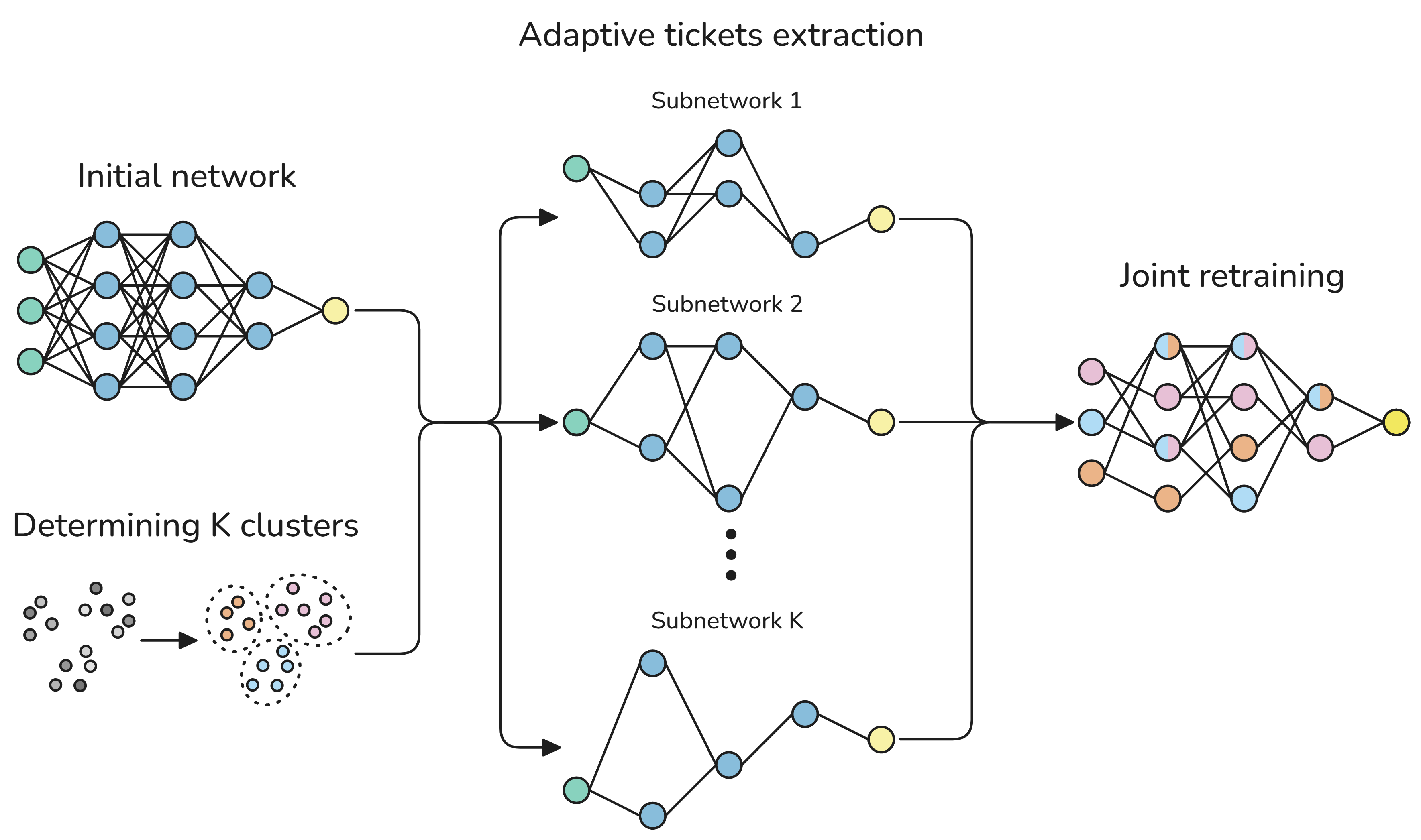}
    \caption{Adaptive pruning pipeline. First, the dataset is divided into subsets via predefined clustering. Then we extracts adaptive tickets, i.e. subnetworks, optimized to specific data cluster, and finally we performed network joint retraining.}
    \label{fig1}
    \vspace{-2.5mm}
\end{figure}

We propose an adaptive pruning framework that extends the LTH to support multiple specialized subnetworks.
An overview of the method is shown in Fig.~\ref{fig1}.
Rather than identifying a single ``winning ticket'', our approach discovers distinct subnetworks, each tailored to a specific data cluster.

Let $f(x; \theta)$ denote a neural network parameterized by $\theta \in \mathbb{R}^d$, $\mathcal{T}$ be a given task, and $\mathcal{D}$ a general dataset.
In standard pruning, a binary mask $m \in \{0,1\}^d$ selects a subnetwork $f_m$ via element-wise multiplication:
\begin{equation}
 f_m = f(x; m \odot \theta),
\end{equation}
\noindent where $\odot$ denotes the Hadamard product.
The LTH posits that there exists a mask $m^\ast$ such that the corresponding subnetwork, when trained in isolation from random initialization, matches the performance of the dense model, i.e.: 
\begin{equation}
    \mathcal{L}_{f_m}(\mathcal{D},\mathcal{T}) = \mathcal{L}_{f}(\mathcal{D},\mathcal{T}),
\end{equation}
\noindent where $\mathcal{L}_{:}$ denotes the model's performance on task $\mathcal{T}$ and dataset $\mathcal{D}$.

While prior work seeks a single universal mask $m^\ast$, we hypothesize that different data subsets $\mathcal{D}_k$ benefit from distinct, pruned subnetworks.
The remainder of this section is organized as follows:  Sec.~\ref{ad_pruning_alg} formalizes the adaptive pruning objective, Sec.~\ref{mask_extraction} describes adaptive mask extraction, and Sec.~\ref{sec:joint_retraining} presents the joint retraining procedure.

\subsection{Formulation of the adaptive pruning algorithm}
\label{ad_pruning_alg}
Given dataset $\mathcal{D}$ and task $\mathcal{T}$, we partition $\mathcal{D}$ into $K$ subsets $\{\mathcal{D}_1, \ldots, \mathcal{D}_K\}$ corresponding to classes or clusters. Such labels can be obtained from supervised labeling systems, such as class annotations, or derived through unsupervised clustering, enabling flexible adaptation across settings.

For each subset $\mathcal{D}_k$, our method involves learning a dedicated mask $m_k$, yielding a subnetwork:
\begin{equation}
\varphi_{m_k} = f(x; m_k \odot \theta), \quad \text{for } x \in \mathcal{D}_k.
\end{equation}
\noindent Given this setup, our objective is to jointly learn the set of masks $ \mathcal{M}^{*} = \{m_1^*, \ldots, m_K^*\}$ that maximize predictive performance on task $\mathcal{T}$ under a sparsity constraint:
\begin{align}
    \mathcal{L}_{\mathcal{M}^{*}} &= \min_{m_k \in \mathcal{M}^{*}} \sum_{k=1}^K \mathbb{E}_{(x,y) \sim \mathcal{D}_k} \left[ \ell\left(y, f\left(x; \, m_k \odot \theta\right)\right) \right]
 \\ 
   &\text{subject to} \quad \|m_k\|_0 \leq s  \quad \forall k,
\end{align} 
\noindent where $\ell$ denotes the loss function and $s$ controls the maximum number of non-zero parameters per subnetwork.

\begin{algorithm}[t]
\caption{Adaptive ticket extraction}\label{alg1}
\small
\begin{algorithmic}
\STATE {\bfseries Input:} Network $f(x; \theta_0)$, train data $\mathcal{D}$, number of subsets $K$, target sparsity $s$, pruning factor $p$, steps $T$, training epoch $N$.
\STATE {\bfseries Output:} Set of $K$ mask $\mathcal{M}$.
\vspace{0.5em}
\STATE  Random initialization of $f(x; \theta_0)$.
\STATE $m_k \gets \mathbf{1}^{|\theta_0|}$ for $k = 1,\ldots,K$.
\STATE $\mathcal{D}_K = \{d_1, ..., d_K\}$ $\gets$ partition of $\mathcal{D}$ into $K$ subsets.
\STATE $\theta \gets \theta_0$

\WHILE{density$(m_K) < s$}
    \FOR{$k = 1$ \textbf{to} $K$}
        \STATE $d_k \gets k$-th subset from $\mathcal{D}_K$
        \STATE  $f(\cdot, \theta_T^{(k)}) \gets $ optimize $f$ for $T$ steps on $d_k$
        \STATE $m_k \gets$ pruning $f$ with factor $p$
        \STATE $\theta \gets \theta_0$.
    \ENDFOR
\ENDWHILE
\end{algorithmic}
\end{algorithm}

\subsection{Adaptive tickets extraction}
\label{mask_extraction}

We now describe how adaptive tickets are extracted for each data subset (see Alg.~\ref{alg1}). Starting from a random initialization $f(x;\theta_0)$, we create an equal-size binary mask $m_k$ for each of the $K$ target subsets, initialized to ones. The dataset is then partitioned into $K$ disjoint subsets  $\mathcal{D}_K = \{d_1, ..., d_K\}$ using a predefined rule, like manual labeling or automatic clustering. Importantly, RTL is fully agnostic with respect to how these subsets are defined, allowing flexibility across diverse tasks and data modalities. 

Each pruning iteration proceeds sequentially over all subsets. For a given subset $d_k$, the network $f(x; \theta_0)$ is trained for $T$ steps to obtain temporary parameters $\theta_T^{(k)}$. We then perform pruning on $f$, removing the lowest-magnitude weights from $\theta_T^{(k)}$ by a fraction $p$. From the set of removed parameters, we yield a sparse subnetwork defined by the mask $m_k$. The remaining weights are subsequently reset to their initial values $\theta_0$. This process repeats until each mask reaches the desired sparsity level $s$, producing the final set of $K$ adaptive masks $\mathcal{M} = \{m_1, \ldots, m_K\}$. Each mask defines a specialized subnetwork $f(x;m_k\odot\theta_0)$.

\begin{algorithm}[t]
\caption{Joint retraining of the adaptive tickets}\label{alg2}
\small
\begin{algorithmic}
\STATE {\bfseries Input:} Network $f(x; \theta)$, train data $\mathcal{D}$, number of subsets $K$, training epochs $N$, learning rate $\eta$, mask set $\mathcal{M}$.
\STATE {\bfseries Output:} $K$ optimized adaptive tickets $f_k(\cdot;\theta_k)$.

\vspace{0.5em}
\STATE \textbf{Balanced dataset creation}
\STATE $\mathcal{D}_K = \{d_1, ..., d_K\}$ $\gets$ partition of $\mathcal{D}$ into $K$ subsets.
\FOR{$k = 1$ \textbf{to} $K$}
    \STATE $d_k \gets$ $k$-th subset from $\mathcal{D}_K$
    \STATE $\mathcal{B}^{k} \gets \{B_1^k,...,B_{M_k}^k \}$
    \COMMENT{Set of batches from $d_k$}
    \STATE $M_k \gets |\mathcal{B}^k|$
\ENDFOR
\STATE $M \gets \max(M_1, \ldots, M_K)$.
\FOR{$k = 1$ \textbf{to} $K$}
    \STATE Repeat batches in $\mathcal{B}^k$ cyclically until $|\mathcal{B}^k| = M$.
\ENDFOR

\vspace{0.5em}
\STATE \textbf{Model training}
\FOR[Training epochs]{$j = 1$ \textbf{to} $N$}
    \FOR[Batch indices]{$m = 1$ \textbf{to} $M$}
        \FOR[Subnetwork indices]{$k = 1$ \textbf{to} $K$}
            \STATE $m_k \gets \mathcal{M}[k]$.
            \STATE $f_k(\cdot; \theta_k) \gets f(\cdot; m_k \odot \theta)$ \COMMENT{$k$-th subnetwork}
            \STATE $(x_m^k, y_m^k) \gets B_m^k$. \COMMENT{Input and label }
            \STATE $\theta_k \leftarrow \theta_k - \eta \left( \nabla_{\theta} 
                \mathcal{L}(f_k(x_m^k; \theta_k), y_m^k) \odot m_k \right)$.
        \ENDFOR
    \ENDFOR
\ENDFOR
\end{algorithmic}
\end{algorithm}

\subsection{Joint Retraining of adaptive tickets}
\label{sec:joint_retraining}

After obtaining the sparse masks $ \mathcal{M} = \{m_1, \dots, m_K\}$, we perform a lightweight \textit{joint retraining phase} to refine subnetwork performance and reinforce specialization. This step, outlined in Alg.~\ref{alg2}, is crucial in our proposed methodology, since it preserves the sparsity structure discovered during mask extraction (alg.~\ref{alg1}) and fine-tunes only the active weights without altering mask topology.

The process begins with balanced dataset creation. As before, the training set $\mathcal{D}$ is partitioned into $K$ subsets $\mathcal{D}_K = \{d_1, \ldots, d_K\}$, each corresponding to a target class or cluster. Each subset $d_k$ is divided into mini-batches $\mathcal{B}^k=\{B_1^k,\ldots,B_{M_k}^k\}$. To ensure synchronized updates across subnetworks, we repeat batches from smaller subsets cyclically until all subsets contain the same number of batches $M=\max(M_1,\ldots,M_K)$. This balancing step guarantees that each subnetwork receives the same number of gradient updates per epoch, even if subsets differ in size.

During joint retraining, each subnetwork $f(x; m_k \odot \theta)$ is trained exclusively on its corresponding data subset $d_k$.
We interleave mini-batches from different subsets and apply gradient updates to the shared dense parameter tensor $\theta$, masking out gradients for pruned weights. Specifically, for subnetwork $k$, the parameter update is given by:
\begin{align}
f_k(\cdot,\theta_k) &\gets f(\cdot, m_k \odot \theta)\\
\theta_k \leftarrow \theta - &\eta \left( \nabla_{\theta} \mathcal{L}(f_k(x^k; \theta_k), y^k) \odot m_k \right) \label{gradient}
\end{align}
\noindent where $\eta$ is the learning rate, $\mathcal{L}$ denotes the empirical loss, while $(x^k,y^k)$ represents the input data with its correspondent label coming from the $k$-th subset. 
By masking the gradient as in Eq.~\ref{gradient}, only the weights retained by mask $m_k$ are updated, preventing interference between subnetworks and avoiding both catastrophic forgetting and collapse. Maintaining limited overlap between subnetworks is crucial for ensuring optimal performance within each cluster without mutual interference or cancellation.

\section{Experiments}
\label{sec:experiments}

We present a sequence of experiments that progressively validate our adaptive pruning framework, moving from controlled settings to real-world applications: (i) class-specific subnetworks on CIFAR-10 \cite{Krizhevsky2009LearningML}, (ii) cluster-aware pruning on CIFAR-100, (iii) implicit neural representations (INRs) with within-image semantic specialization, (iv) speech enhancement in heterogeneous acoustic environments, and (v) subnetwork overlap and semantic alignment analysis.

Across all tasks, we control for architecture, sparsity, and training budget to isolate the effects of specialization. Quantitative results and discussion are reported in Sec.~\ref{sec:results}, while full implementation details are provided in the appendix.

\subsection{CIFAR-10 subnetwork specialization}

We begin with a controlled CIFAR-10 experiment, where each of the 10 classes defines a distinct data subset and is assigned its own subnetwork. This idealized setting allows us to directly test the core premise of our approach: whether subnetworks specialized to disjoint subsets outperform a single universal pruning mask under identical constraints.

We compare \textit{RTL} against two baselines:  
(i) IMP (single model), which produces a single shared subnetwork across all classes, and  
(ii) IMP (multiple models), which independently prunes one model per class without weight sharing.

All methods use the same backbone architecture, sparsity targets, and pruning budget. Performance is evaluated using balanced accuracy, precision, recall, and parameter count at 25\%, 50\%, and 75\% sparsity. This experiment establishes an upper bound on specialization benefits and serves as a reference point for more challenging scenarios. Training protocols and architectural details are provided in Appendix~\ref{app:vision_setup}.

\subsection{Cluster-aware pruning on CIFAR-100}

To assess robustness under less ideal conditions, we evaluate RTL on CIFAR-100, where class boundaries are fine-grained and semantically overlapping. Instead of assigning subnetworks to individual classes, we group the 100 classes into 8 coarse semantic clusters using an unsupervised text-based clustering procedure.

The resulting clusters are semantically coherent but imperfectly aligned with visual features, introducing ambiguity that better reflects real-world data partitioning. RTL is compared against the same two IMP baselines under matched conditions, using the same metrics and sparsity levels as in the CIFAR-10 experiment. Details of the clustering pipeline are provided in Appendix~\ref{app:clustering}.

\renewcommand{\arraystretch}{0.5}

\begin{table*}[b]
\small
    \centering
    \caption{Results on CIFAR-10 with class-specific subnetworks and CIFAR-100 with cluster-specialized subnetworks. We report balanced accuracy, precision, recall, and the number of remaining parameters at three sparsity levels. RTL is compared against (a) IMP with a single shared subnetwork and (b) IMP trained independently per class (multiple models). Best values per metric and sparsity level are bolded.  Clusters are derived from semantic embeddings (see Section~\ref{sec:experiments}).}
    \begin{tabular}{@{}cc|cccccccccccc@{}}
         & \multirow{2}{*}{\textbf{Method}} & \multicolumn{3}{c}{\textbf{Balanced accuracy $\uparrow$}} & \multicolumn{3}{c}{\textbf{Precision $\uparrow$}} & \multicolumn{3}{c}{\textbf{Recall $\uparrow$}} & \multicolumn{3}{c}{\textbf{\#Params $\downarrow$}} \Tstrut\Bstrut\\[0.1cm]
         \cmidrule(lr){3-5} \cmidrule(lr){6-8} \cmidrule(lr){9-11} \cmidrule(lr){12-14}
         & & 25\% & 50\% & 75\% & 25\% & 50\% & 75\% & 25\% & 50\% & 75\% & 25\% & 50\% & 75\% \Tstrut\Bstrut\\[0.1cm]
         \hline
         \multirow{3}{*}{\rotatebox[origin=c]{90}{\parbox[c]{1.7cm}{\centering CIFAR-10}}}& RTL (ours) & \textbf{0.781} & \textbf{0.778} & \textbf{0.772} & 0.282 & 0.276 & 0.257 & \textbf{0.821} & \textbf{0.810} & \textbf{0.816} & 103K & 72K & 38K \TTstrut\\[0.25cm]
         & \makecell{IMP\\(single model)} & 0.711 & 0.711 & 0.732 & \textbf{0.479} & \textbf{0.478} & \textbf{0.515} & 0.480 & 0.480 & 0.518 & \textbf{94K} & \textbf{63K} & \textbf{31K} \\[0.3cm]
         & \makecell{IMP\\(multiple models)} & 0.712 & 0.710 & 0.760 & 0.233 & 0.222 & 0.262 & 0.701 & 0.702 & 0.766 & 944K & 629K & 314K \\[0.3cm]
         \hline
         \multirow{3}{*}{\rotatebox[origin=c]{90}{\parbox[c]{1.8cm}{\centering CIFAR-100}}}& RTL (ours) & \textbf{0.765} & \textbf{0.751} & \textbf{0.759} & 0.298 & 0.290 & 0.289 & \textbf{0.764} & \textbf{0.729} & \textbf{0.754} & 108K & 76K & 40K \TTstrut\\[0.25cm]
         & \makecell{IMP\\(single model)} & 0.722 & 0.707 & 0.742 & \textbf{0.45} & \textbf{0.463} & \textbf{0.522} & 0.42 & 0.421 & 0.463 & \textbf{94K} & \textbf{63K} & \textbf{31K} \\[0.3cm]
         & \makecell{IMP\\(multiple models)} & 0.712 & 0.700 & 0.744 & 0.276 & 0.271 & 0.286 & 0.660 & 0.637 & 0.732 & 944K & 629K & 314K \\[0.3cm]

    \end{tabular}
    \label{tab:cifar10_and_cifar100}
\end{table*}

\subsection{Implicit Neural Representations }

We evaluate RTL in the context of INRs, where specialization is defined over semantic regions within a single image. The task consists of reconstructing an image by mapping continuous pixel coordinates to RGB values using a coordinate-based neural network.

Experiments are conducted on 10 images from the ADE20K  dataset \cite{zhou2019semantic} selected to ensure diversity in scene category and luminosity/color characteristics. Semantic segmentation masks define region-level classes, and RTL learns specialized subnetworks for these regions, while a standard baseline conditions a single network on region identity via class embeddings.

Reconstruction quality is evaluated using peak signal-to-noise ratio (PSNR), averaged over all pixels and all images. Architectural choices, positional encoding, training protocol, and per-image results are reported in Appendix~\ref{app:inr}.

\subsection{Speech enhancement in realistic environments}

We evaluate RTL on a real-world speech enhancement task characterized by heterogeneous acoustic conditions. Clean speech is mixed with noise from three distinct acoustic scenes (\textit{indoor}, \textit{outdoor}, and \textit{transportation}), each defining a specialization subset.

RTL learns one subnetwork per acoustic scene and is compared against (i) a single IMP-pruned model shared across all environments and (ii) independently pruned IMP models without weight sharing. All methods operate under identical sparsity and computational constraints. Performance is measured using scale-invariant signal-to-noise ratio improvement (SI-SNRi). Dataset construction, model architecture, and signal processing details are provided in Appendix~\ref{app:speech_setup}.

\subsection{Subnetwork collapse and semantic alignment}

Beyond task-level performance, we analyze relationships among learned subnetworks in the CIFAR-10 and CIFAR-100 experiments. Specifically, we study how subnetwork overlap evolves with increasing sparsity and how excessive overlap relates to performance degradation, a phenomenon we refer to as \emph{subnetwork collapse}.

We quantify pairwise mask similarity at multiple sparsity levels and examine its correlation with balanced accuracy. Finally, using CIFAR-10, we compare structural similarity with semantic distances between class labels derived from WordNet, assessing whether pruning structures encode high-level conceptual relationships. Formal definitions and analysis procedures are provided in Appendix~\ref{app:analysis}.

\section{Results}
\label{sec:results}

\subsection{CIFAR-10 subnetwork specialization}

Table~\ref{tab:cifar10_and_cifar100} summarizes CIFAR-10 results under class-specific pruning. RTL consistently achieves the highest balanced accuracy and recall across all sparsity levels. At 25\% sparsity, RTL attains a balanced accuracy of 0.781, significantly outperforming both baselines (0.711 for single-model IMP and 0.712 for multi-model IMP). This advantage persists at 50\% sparsity (0.778 vs. 0.711) and remains competitive even at 75\% sparsity (0.772 vs. 0.760 for multi-model IMP), despite using an order of magnitude fewer parameters.

RTL also achieves the highest recall at all sparsity levels (0.821 / 0.810 / 0.816), exceeding the single-model baseline and matching or surpassing the multi-model alternative. This demonstrates that RTL subnetworks effectively preserve class-relevant signals under aggressive pruning.

The lower precision of RTL (0.257-0.282) compared to the single-model IMP baseline (0.478-0.515) reflects a design trade-off: by prioritizing recall, RTL subnetworks favor sensitivity to true positives over strict discrimination, which is well-suited for class-specialized inference. In downstream applications, precision can be recovered via thresholding or ensemble methods if needed.

Critically, these gains come with exceptional parameter efficiency. RTL uses only 103K / 72K / 38K parameters at 25\% / 50\% / 75\% sparsity, compared to 944K / 629K / 314K for multi-model IMP (nearly 10× more) to achieve comparable accuracy. This underscores RTL's ability to deliver high performance without sacrificing compactness.

\begin{figure*}[b]
    \vspace{-4mm}
    \centering
    \includegraphics[width=0.9\linewidth]{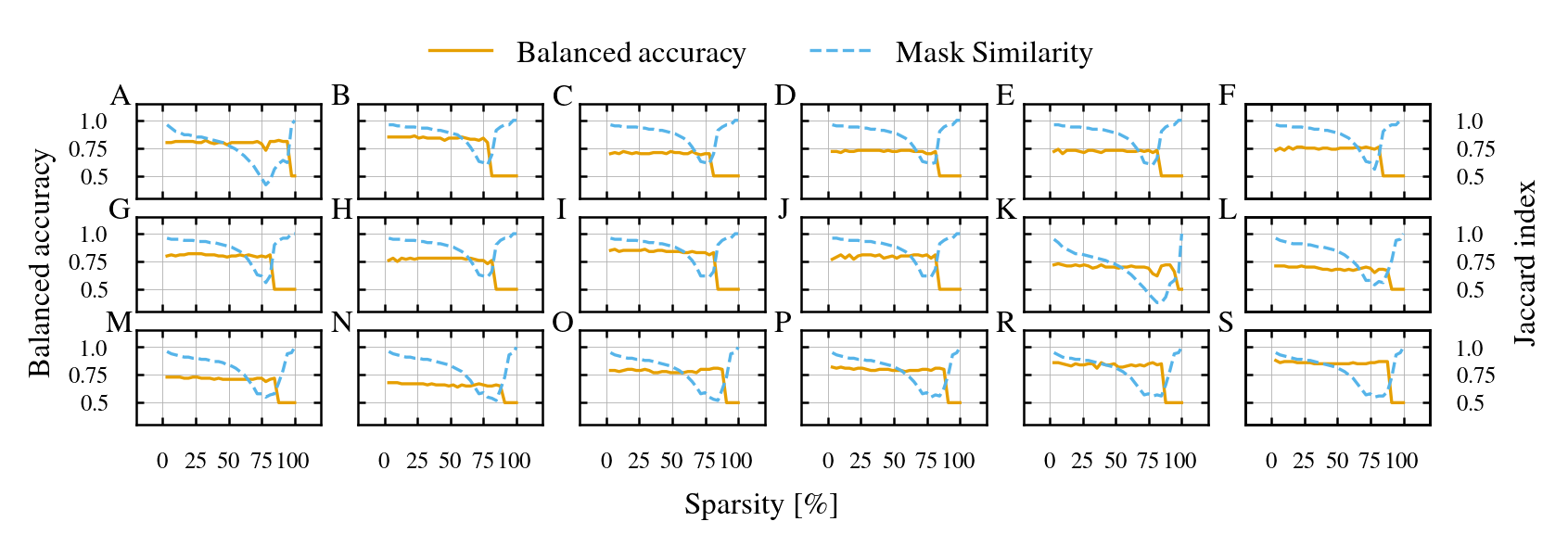}
    \caption{Mask collapse analysis on CIFAR-10 and CIFAR-100. Each subplot corresponds to one class and shows balanced accuracy (solid line) and mask similarity to other subnetworks (dashed line). Plots A-J corresponds to CIFAR-10 network and plots K-S to CIFAR-100.}
    \label{fig:cifar10_cifar100_collapse}
\end{figure*}

\subsection{Cluster-aware pruning on CIFAR-100}

Table~\ref{tab:cifar10_and_cifar100} also reports CIFAR-100 results, where subnetworks are specialized to semantically derived clusters. Despite the inherent ambiguity in this partitioning, RTL again achieves the highest balanced accuracy across all sparsity levels: 0.765 (25\%), 0.751 (50\%), and 0.759 (75\%). Notably, RTL outperforms both baselines even at 75\% sparsity, where model capacity is most constrained, demonstrating robustness to imperfect data grouping.

Recall follows the same trend: RTL achieves 0.764 / 0.729 / 0.754, substantially exceeding the single-model IMP baseline (0.420-0.463) and consistently outperforming the multi-model alternative. This confirms that RTL subnetworks retain cluster-discriminative features more effectively under pruning, even when cluster boundaries are noisy.

As in CIFAR-10, RTL exhibits lower precision than the single-model baseline, reflecting its emphasis on sensitivity over selectivity. Given the coarse and overlapping nature of the clusters, high recall is particularly valuable for ensuring coverage of relevant visual concepts.

In terms of efficiency, RTL uses only 108K / 76K / 40K parameters, comparable to the single-model model baseline and drastically fewer than the multi-model baseline (944K / 629K / 314K). This reaffirms that RTL achieves strong specialization without requiring redundant, over-parameterized subnetworks, making it especially suitable for scenarios with limited memory or compute.

\subsection{Implicit Neural Representations}

\begin{table}[t]
\footnotesize
    \centering
    \caption{INR results on ADE20k dataset. Methods are evaluated at 25\%, 50\%, and 75\% sparsity using PSNR, reporting the number of trainable parameters.}
    \begin{tabular}{c|c@{\hskip 2mm}c@{\hskip 2mm}cc@{\hskip 2mm}c@{\hskip 2mm}c}
         \multirow{2}{*}{\textbf{Method}} & \multicolumn{3}{c}{\textbf{PSNR $\uparrow$}} & \multicolumn{3}{c}{\textbf{\#Params $\downarrow$}} \Tstrut\Bstrut\\[0.1cm]
         \cmidrule(lr){2-4} \cmidrule(lr){5-7}
         & 25\% & 50\% & 75\% & 25\% & 50\% & 75\% \Tstrut\Bstrut\\[0.1cm]
         \hline
         RTL (ours)                         & \textbf{18.86} & \textbf{17.25} & \textbf{14.87} & 48.0K          & 36.5K          & 22.0K \TTstrut \\[0.25cm]
         \makecell{IMP\\(single model)}     & 15.94          & 14.72          & 12.69          & \textbf{40.5K} & \textbf{27.0K} & \textbf{13.5K} \\[0.3cm]
    \end{tabular}
    \label{tab:INR}
    \vspace{-2mm}
\end{table}

Table~\ref{tab:INR} reports reconstruction performance for the INR experiment on ADE20K images at varying sparsity levels.

Across all sparsity regimes, RTL consistently outperforms the standard IMP baseline. At 25\% sparsity, RTL achieves a PSNR of 18.58, exceeding the single-model IMP baseline by almost 3~dB. This performance gap remains substantial as sparsity increases, with RTL maintaining advantages of 2.53~dB and 2.18~dB at 50\% and 75\% sparsity, respectively.

Notably, these gains are achieved despite RTL retaining a larger number of parameters than the single-model IMP baseline. This behavior mirrors observations from the classification and speech enhancement experiments: enforcing a single global pruning mask across heterogeneous subsets (in this case, semantically distinct image regions) leads to suboptimal allocation of model capacity. By contrast, RTL enables region-specific subnetworks that preserve functionally important parameters, resulting in higher reconstruction fidelity under identical computational budgets.

As sparsity increases, performance degrades for both methods, but RTL degrades more gracefully, indicating that specialization is especially beneficial in high-sparsity regimes, where competition for shared parameters intensifies. The results demonstrate that adaptive pruning can effectively capture semantic structure even in coordinate-based representations, extending the benefits of specialization beyond dataset-level tasks to within-image semantic decomposition.

Per-image PSNR values are reported in Appendix~\ref{app:inr}, confirming that the observed trends are consistent across images with diverse scene content and appearance characteristics.

\subsection{Speech Enhancement in realistic environments}

\begin{table}[t]
\footnotesize
    \centering
    \caption{Speech enhancement results on  DNS Challenge 2020 and TAU Urban Acoustic Scenes 2020 datasets. Methods are evaluated at 25\%, 50\%, and 75\% sparsity using SI-SNRi, reporting the number of trainable parameters.}
    \begin{tabular}{c|c@{\hskip 2mm}c@{\hskip 2mm}cc@{\hskip 2mm}c@{\hskip 2mm}c}
         \multirow{2}{*}{\textbf{Method}} & \multicolumn{3}{c}{\textbf{SI-SNRi $\uparrow$}} & \multicolumn{3}{c}{\textbf{\#Params $\downarrow$}} \Tstrut\Bstrut\\[0.1cm]
         \cmidrule(lr){2-4} \cmidrule(lr){5-7}
         & 25\% & 50\% & 75\% & 25\% & 50\% & 75\% \Tstrut\Bstrut\\[0.1cm]
         \hline
         RTL (ours) & \textbf{7.248} & \textbf{7.178} & \textbf{6.992} &  32.0K & 22.8K & 12.3K \TTstrut\\[0.25cm]
         \makecell{IMP\\(single model)} & 6.885 & 6.97 & 6.967 & \textbf{28.0K} & \textbf{18.6K} & \textbf{9.3K} \\[0.3cm]
         \makecell{IMP\\(multiple model)} & 5.295 & 5.775 & 5.899 & 84.1K & 56.1K & 28.0K \\[0.3cm]

    \end{tabular}
    \label{tab:SE}
    \vspace{-2mm}
\end{table}

\begin{figure*}[t]
    \centering
    \includegraphics[width=0.90\linewidth]{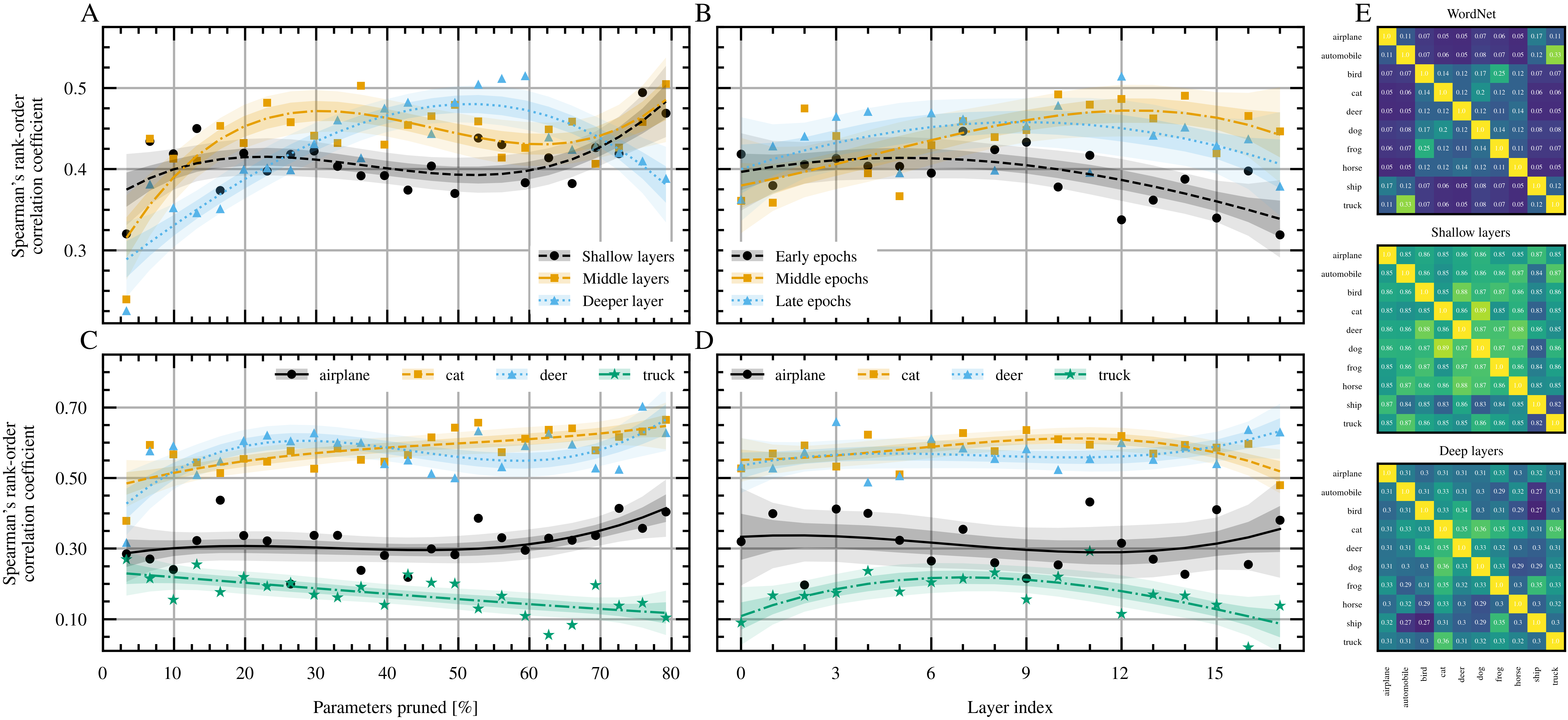}
    \caption{Semantic and structural correlation analysis. (A) Spearman's rank-order correlation between semantic similarity and mask similarity versus pruning ratio across shallow, middle, and deep layers. (B) Correlation across depth for early, middle, and late training stages. (C) Correlation versus pruning ratio for four representative classes: \textit{airplane}, \textit{cat}, \textit{deer}, and \textit{truck}. (D) Correlation across depth for the same four classes. (E) WordNet path similarity (top) and RTL mask similarity matrices for shallow (middle) and deep (bottom) layers.}
    \label{fig:semantic_sim}
\end{figure*}

Table~\ref{tab:SE} reports speech enhancement performance across three acoustic environments (indoor, outdoor, and transportation). RTL achieves the highest SI-SNRi at all sparsity levels: 7.248, 7.178, and 6.992 at 25\%, 50\%, and 75\% sparsity, respectively, consistently outperforming both IMP baselines. This demonstrates that environment-specific subnetworks better capture each noise type's distinct spectro-temporal characteristics, yielding superior waveform reconstruction.

The single-mask IMP model performs reasonably well, likely by learning a general-purpose denoising strategy. However, it lags behind RTL, confirming that specialization yields tangible gains even in a non-classification task. In contrast, the multi-model IMP baseline underperforms despite a larger parameter budget (84.1K vs. 32.0K at 25\% sparsity), suggesting that independent pruning without shared initialization or joint retraining yields suboptimal results.

Critically, RTL achieves these gains with only modest overhead in model size: 32.0K / 22.8K / 12.3K parameters, just slightly larger than the single-mask model and less than half the size of the multi-model alternative. Notably, even the most aggressively pruned RTL subnetwork (12.3K parameters) surpasses both baselines in SI-SNRi, underscoring the efficiency of our approach.

These results confirm that RTL generalizes beyond controlled vision benchmarks: when subnetworks align with meaningful real-world structure (in this case, acoustic environments), they deliver significantly better performance while remaining compact and deployable.

\subsection{Subnetworks collapse and semantic alignment}

\paragraph{Subnetworks collapse} 
Fig.~\ref{fig:cifar10_cifar100_collapse} illustrates the relationship between performance and mask similarity across sparsity levels for CIFAR-10 and CIFAR-100. For each class or cluster, we plot (i) balanced accuracy and (ii) average Jaccard similarity (IoU) to all other subnetworks. These curves reveal the onset of subnetwork collapse - a failure mode in which excessive pruning forces subnetworks to converge to overlapping weight sets, eroding their specialization.

In CIFAR-10 (Fig.~\ref{fig:cifar10_cifar100_collapse}A-J), where classes are perfectly separated, most subnetworks maintain high accuracy and low mask similarity up to 70-80\% sparsity. Beyond this point, a sharp increase in IoU coincides with a precipitous drop in accuracy, confirming that performance degradation is directly linked to loss of structural distinctiveness. Crucially, the IoU spike reliably precedes or coincides with the accuracy drop, suggesting that mask similarity can serve as a label-free early-warning signal for oversparsification.

On CIFAR-100 (Fig.~\ref{fig:cifar10_cifar100_collapse}K-S), where classes are grouped into eight semantically derived clusters, the same collapse signature emerges: accuracy remains stable as long as mask similarity is bounded, but deteriorates sharply once pruning forces subnetworks into excessive overlap. Collapse occurs at slightly higher sparsity thresholds than in CIFAR-10, likely because clusters share more visual features, allowing for greater weight reuse before specialization is lost.

Across both datasets, the evidence is consistent: subnetwork specialization is essential for RTL's performance, and mask similarity is a robust, label-free predictor of collapse. Moreover, the abruptness of the accuracy drop suggests that once specialized weights are pruned beyond a critical point, recovery is unlikely without full retraining, highlighting the importance of stopping pruning before collapse occurs.

\paragraph{Semantic alignment}
Fig.~\ref{fig:semantic_sim} summarizes how semantic alignment relates to subnetwork structure throughout training and across depth. Panels A-D report Spearman's rank-order correlations between semantic proximity (from WordNet) and mask similarity (Jaccard index), while panel E shows the corresponding WordNet and mask similarity matrices for shallow and deep layers.

Panel A shows pruning correlations across layer depths: shallow and middle layers increase correlation up to 20-35\% sparsity, then decline, indicating that moderate pruning enhances semantic organization in early layers, while excessive sparsity degrades it. Deep layers start with low correlations but rise steadily to around 55\% sparsity, reflecting stronger semantic alignment as higher-level representations develop.

Panel B tracks training evolution: early optimization shows weak correlations decreasing with depth, but later epochs reveal consistent increases, particularly in middle and deep layers, indicating a gradual emergence of semantic structure as subnetworks refine specialized connectivity.

Panels C and D highlight four representative classes (\textit{airplane}, \textit{cat}, \textit{deer}, \textit{truck}). Subnetworks for semantically related classes (\textit{cat}, \textit{deer}) show stronger correlations (0.6-0.7) that grow with pruning and depth, while unrelated ones (\textit{airplane}, \textit{truck}) remain weaker (0.2-0.3).
Thus, RTL subnetworks for related categories preserve overlapping structures, whereas distant ones evolve independently - a pattern consistent with overall mask stability across dissimilar classes.

Panel E provides complementary heatmaps. The top matrix shows WordNet path similarities, followed by mask similarities for shallow and deep layers. Shallow masks are uniformly similar (mean $\approx$ 0.86), reflecting shared early filters for low-level, class-agnostic features such as edges and textures. Slight local increases for related pairs, for example \textit{cat-dog} (0.89), \textit{automobile-truck} (0.87), and \textit{deer-horse} (0.88), suggest broad semantic grouping even at early stages. In deep layers, mask similarity drops (mean $\approx$ 0.31), and a clear block-diagonal structure emerges that aligns with WordNet relations. Related animal classes (\textit{cat}, \textit{dog}, \textit{horse}, \textit{deer}) show higher overlap (0.33-0.36) than unrelated ones (0.27-0.31). RTL forms weaker specialization among vehicle classes (\textit{airplane}, \textit{automobile}, \textit{ship}, \textit{truck}), likely due to limited visual similarity beyond the \textit{automobile-truck} pair.

Overall, RTL subnetworks gradually organize according to semantic structure in the data. Early layers are shared and class-agnostic, while deeper layers increasingly reflect conceptual hierarchies. The growing correlation with both depth and training indicates that RTL pruning not only enforces sparsity but also promotes semantically meaningful specialization within a shared model architecture.

\section{Conclusion}
\label{sec:conclusion}

We introduced Routing the Lottery (RTL), an adaptive pruning framework that discovers multiple specialized subnetworks (\textit{adaptive tickets}) rather than a single universal winning ticket. Across multiple settings RTL consistently achieves superior or competitive performance while maintaining a compact parameter footprint. These results show that specialization naturally emerges when subnetworks are allowed to diverge in response to data heterogeneity.

Our analysis reveals two critical insights. First, subnetwork distinctiveness is essential: mask similarity serves as a reliable, label-free indicator of subnetwork collapse, showing that the bottleneck is not raw capacity, but the preservation of structural diversity.
Second, RTL is robust to imperfect data partitions - it excels with clean class boundaries (CIFAR-10), remains effective under noisy semantic clustering (CIFAR-100), and generalizes to real-world applications where subnetworks align with meaningful environmental factors. 

We also observe that RTL subnetworks tend to favor recall over precision, increasing sensitivity to weak class- or environment-specific signals. While beneficial for routing-based inference, this suggests that downstream calibration or selective filtering could further improve precision without compromising specialization.

This work positions adaptive pruning as a pathway toward more modular, efficient, and interpretable deep models, ones that dynamically allocate representational capacity in alignment with the intrinsic structure of the data they process.

\section*{Impact Statement}

This paper presents work whose goal is to advance the field of Machine
Learning. There are many potential societal consequences of our work, none
which we feel must be specifically highlighted here.


\bibliography{rtl}
\bibliographystyle{icml2026}

\newpage
\appendix
\onecolumn
\section{Vision Model and Training Setup}
\label{app:vision_setup}

\subsection{Model architecture and pruning scope}

For all computer vision experiments on CIFAR-10 and CIFAR-100, we use GhostNet \cite{Han2019} as the backbone. To control model capacity and isolate the effects of adaptive pruning, we retain the convolutional stem followed by the first nine Ghost bottleneck blocks, discarding deeper stages of the original network.

All convolutional layers are implemented as masked convolutions and constitute the pruning scope. Batch normalization layers, squeeze-and-excitation modules, and the final classifier remain dense. Under this configuration, the model contains approximately 126K prunable parameters. 

The retained GhostNet variant follows the standard Ghost bottleneck design, combining inexpensive depthwise convolutions with pointwise expansions and optional downsampling and squeeze-and-excitation. Spatial resolution is progressively reduced via strided depthwise convolutions in selected bottlenecks. A detailed summary of the architecture, including channel dimensions and downsampling stages, is provided in Table~\ref{tab:cifar_ghostnet_arch} to ensure reproducibility.

\begingroup
\setlength{\tabcolsep}{10pt}
\renewcommand{\arraystretch}{1.5}
\begin{table}[h]
\centering
\small
\caption{GhostNet backbone used for CIFAR experiments. Only convolutional layers are pruned. All convolutions are followed by batch normalization and ReLU unless stated otherwise.}
\label{tab:cifar_ghostnet_arch}
\begin{tabular}{c|c|c|c|c}
\hline
\textbf{Stage} & \textbf{Block type} & \textbf{Input $\rightarrow$ Output} & \textbf{Kernel / Stride} & \textbf{Notes} \\
\hline
Stem & Conv2D & $3 \rightarrow 16$ & $3\times3$ / $2$ & Initial downsampling \\
\hline
1 & GhostBottleneck & $16 \rightarrow 16$ & $1\times1$, $3\times3$ / $1$ & No shortcut \\
2 & GhostBottleneck & $16 \rightarrow 24$ & $3\times3$ / $2$ & Downsampling + shortcut \\
3 & GhostBottleneck & $24 \rightarrow 24$ & $1\times1$, $3\times3$ / $1$ & Shortcut \\
4 & GhostBottleneck & $24 \rightarrow 40$ & $5\times5$ / $2$ & SE + shortcut \\
5 & GhostBottleneck & $40 \rightarrow 40$ & $1\times1$, $3\times3$ / $1$ & SE \\
6 & GhostBottleneck & $40 \rightarrow 80$ & $3\times3$ / $2$ & Downsampling + shortcut \\
7 & GhostBottleneck $\times 3$ & $80 \rightarrow 80$ & $1\times1$, $3\times3$ / $1$ & Repeated blocks \\
8 & Conv1$\times$1 & $80 \rightarrow 184$ & $1\times1$ / $1$ & Channel expansion \\
\hline
Head & Conv + Pool & $184 \rightarrow 128$ & $1\times1$ & Global avg pooling \\
Classifier & Linear & $128 \rightarrow C$ & -- & $C=10$ or $100$ \\
\hline
\end{tabular}
\end{table}
\endgroup

\subsection{Optimization and training protocol}

All models are trained using the Adam optimizer \cite{kingma2015adam} with a learning rate of $1\mathrm{e}{-4}$ and no weight decay. Each pruning iteration consists of two phases: (i) pruning and (ii) joint retraining, both run for 10 epochs. Batch sizes are set to 320 for CIFAR-10 and 256 for CIFAR-100.

To ensure fair comparison, all methods (RTL, single-model IMP, and multi-model IMP) share the same random Kaiming initialization \cite{he2015delving}. RTL and multi-model IMP subnetworks are trained as binary classifiers with balanced mini-batches, effectively doubling the number of sample presentations per epoch. To match total data exposure, the single-model IMP baseline is trained for 20 epochs.

\subsection{Pruning schedule and compute}

Pruning follows a fixed schedule, removing 4,096 weights per epoch for each subnetwork until the target sparsity is reached. All experiments are executed on a single NVIDIA H100 GPU. Full RTL and multi-model IMP runs require approximately 6 hours, while the single-model IMP baseline completes in roughly 45 minutes.

\section{Semantic Clustering Procedure}
\label{app:clustering}

To construct coarse semantic subsets for CIFAR-100, we apply an unsupervised clustering pipeline based on textual class descriptions rather than visual features.

First, we extract text embeddings for each of the 100 CIFAR-100 class names using the CLIP text encoder. These embeddings are then reduced in dimensionality using UMAP to facilitate clustering in a lower-dimensional space. Finally, we apply HDBSCAN to the reduced embeddings to obtain cluster assignments.

This procedure yields 8 semantic clusters. While the resulting clusters are semantically coherent, they are not perfectly aligned with visual similarity, intentionally introducing noise and overlap. This imperfect alignment better reflects real-world scenarios where data partitions are ambiguous or only approximately defined. The full cluster composition is reported in Table~\ref{tab:cifar100_clusters}.

\begingroup
\setlength{\tabcolsep}{10pt}
\renewcommand{\arraystretch}{1.5}
\begin{table}[h]
\centering
\small
\caption{Semantic clusters used in the CIFAR-100 experiments. Each cluster defines a data subset to which a specialized subnetwork is assigned.}
\label{tab:cifar100_clusters}
\begin{tabular}{c|p{11cm}}
\hline
\textbf{Cluster} & \textbf{CIFAR-100 Classes} \\
\hline
1 (Aquatic animals) & aquarium fish, dinosaur, dolphin, flatfish, ray, seal, shark, trout, whale \\
\hline
2 (People \& objects) & baby, bed, bicycle, bottle, bowl, boy, bridge, can, castle, chair, clock, couch, cup, girl, house, keyboard, lamp, man, plain, plate, road, rocket, skyscraper, table, telephone, television, wardrobe, woman \\
\hline
3 (Mammals) & bear, beaver, camel, cattle, chimpanzee, elephant, fox, hamster, kangaroo, leopard, lion, mouse, otter, porcupine, possum, rabbit, raccoon, shrew, skunk, squirrel, tiger, wolf \\
\hline
4 (Reptiles \& insects) & bee, beetle, butterfly, caterpillar, cockroach, crab, crocodile, lizard, lobster, mushroom, snail, snake, spider, turtle, worm \\
\hline
5 (Vehicles) & bus, lawn mower, motorcycle, pickup truck, streetcar, tank, tractor, train \\
\hline
6 (Natural scenes) & cloud, forest, maple tree, mountain, oak tree, palm tree, pine tree, sea, willow tree \\
\hline
7 (Fruits) & apple, orange, pear, sweet pepper \\
\hline
8 (Flowers) & orchid, poppy, rose, sunflower, tulip \\
\hline
\end{tabular}
\end{table}
\endgroup

All clustering hyperparameters are fixed across experiments and are not tuned to downstream task performance. As shown in Sec.~\ref{sec:results}, RTL remains effective under this setting, demonstrating robustness to non-ideal specialization boundaries and highlighting the benefits of adaptive pruning beyond strictly class-aligned scenarios.
\section{Implicit Neural Representation}
\label{app:inr}

\subsection{Task setup}

We consider the standard INR formulation of mapping continuous pixel coordinates $(x, y)$ to RGB values. The full image is reconstructed through point-wise evaluation of the network over all pixel locations.

\begin{figure}[h]
    \centering
    \includegraphics[width=0.8\columnwidth]{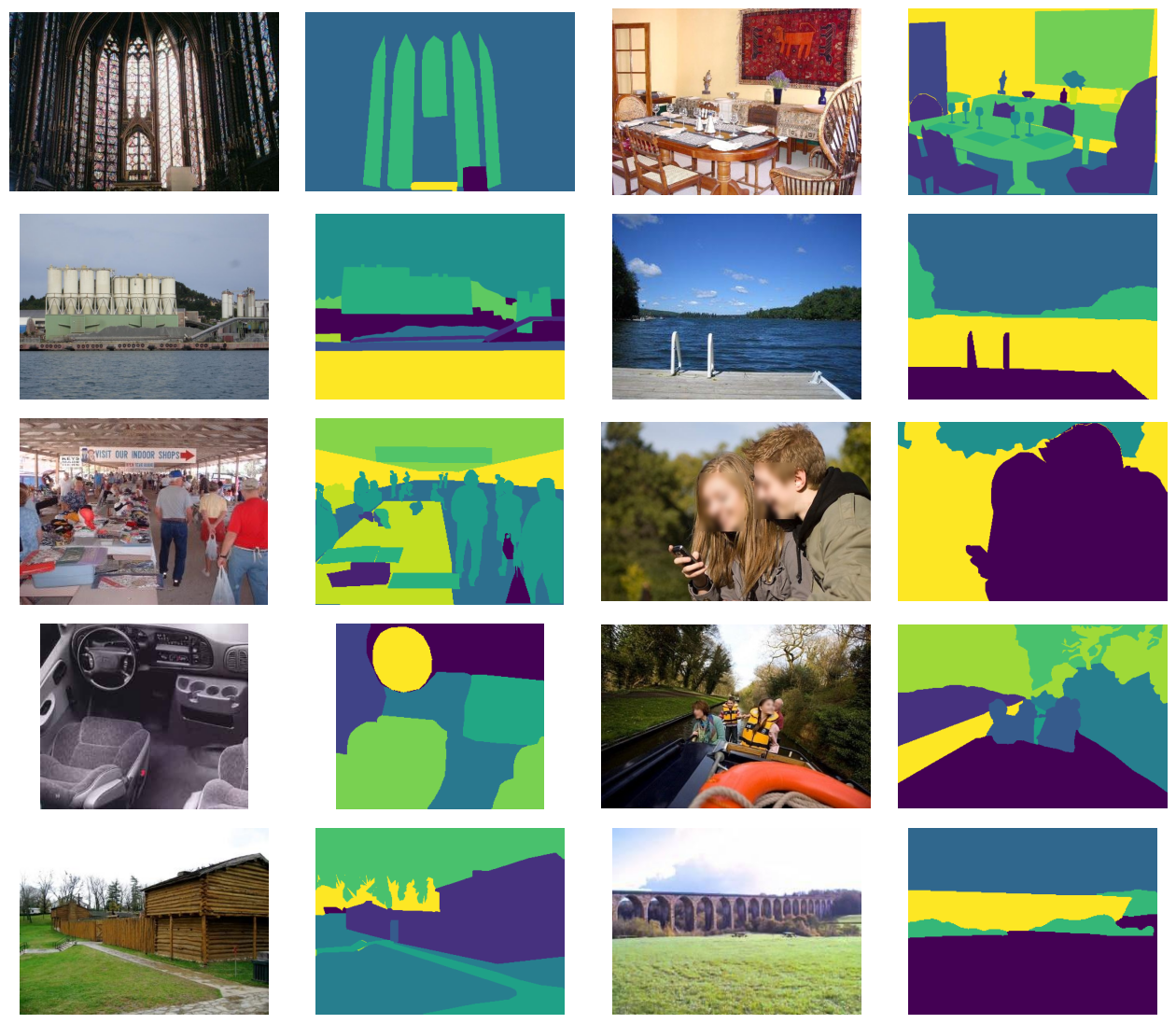}
    \caption{Data samples from the ADE20K dataset used in the INR experiments, shown together with their corresponding preprocessed semantic segmentation masks.}
    \label{fig:ade_samples}
\end{figure}

Experiments are conducted on 10 images selected from the training split of the ADE20K dataset to ensure diversity in scene category and luminosity/color characteristics. Specifically, we use images with indices 0, 3733, 7304, 8399, 11708, 12963, 13783, 14236, 18813, and 23790. Each image is treated independently, and a separate model is trained per image. The selected images and their corresponding preprocessed semantic segmentation masks are shown in Fig.~\ref{fig:ade_samples}.

Semantic segmentation annotations are used to define classes: all masks corresponding to the same semantic object category within an image are merged into a single class (e.g., multiple instances of \emph{tree} are treated as one mask).

The dataset contains small unassigned or noisy regions, typically at object boundaries. These regions are reassigned to a neighboring semantic class, which was empirically found to stabilize training for both RTL and baseline models.

\subsection{Positional encoding}

Input coordinates $(x, y)$ are encoded using Fourier features prior to being passed to the network, following standard INR practice. This encoded representation forms the sole input to the RTL model.

\subsection{Model architectures}

All models are multilayer perceptrons (MLPs) with ReLU activations. For RTL, the network consists of five layers with the following input–output dimensions: $[[128, 34], [128, 128], [128, 128], [128, 128], [3, 128]]$

For the standard baseline, the input is augmented with a learned class embedding corresponding to the semantic mask of each pixel, increasing the first-layer dimensionality to $[128, 50]$. All subsequent layers are identical to those used in RTL.

\subsection{Training and pruning protocol}

All models are trained using the Adam optimizer with a learning rate of 0.01. Training proceeds for 10,000 optimization steps, after which pruning is applied by removing 4,096 weights per pruning iteration.

Following each pruning step, all remaining (non-pruned) weights are rewound to their initial values, consistent with the IMP-style pruning protocol used throughout the paper. This prune-and-rewind process is repeated until the target sparsity is reached.

\subsection{Initialization and fairness considerations}

Because the RTL and baseline models differ in input dimensionality, care is taken to ensure fair initialization. We initialize the larger baseline network (including the class embedding input) and manually remove the weights corresponding to the class embedding dimensions to obtain the RTL initialization. This ensures that all shared parameters are initialized identically across methods.

The same initialization seed is used across all images.

\subsection{Evaluation protocol}

Reconstruction quality is evaluated using peak signal-to-noise ratio (PSNR). In the main text, PSNR is reported as a single scalar value averaged over all pixels and all images. For completeness, we additionally report per-image PSNR values in the appendix to illustrate variability across samples.

\subsection{Per-Image INR Results}
\label{app:inr_per_image}

Table~\ref{tab:inr_per_image} reports per-image reconstruction performance for the INR experiment on 10 ADE20K images. Results are shown for RTL and the single-model IMP baseline at 25\%, 50\%, and 75\% sparsity, measured using PSNR.

\begin{table}[h]
    \centering
    \caption{Per-image PSNR for the INR experiment on 10 ADE20K images. Results are reported for RTL and the single-model IMP baseline at 25\%, 50\%, and 75\% sparsity.}
    \label{tab:inr_per_image}
    \begin{tabular}{c|ccc|ccc}

    & \multicolumn{6}{c}{\textbf{PSNR $\uparrow$}} \Tstrut\Bstrut\\[0.1cm] 
    \textbf{Sample} & \multicolumn{3}{c|}{\textbf{RTL}}
    & \multicolumn{3}{c}{\textbf{IMP (single model)}} \Tstrut\Bstrut\\[0.1cm]
    & 25\% & 50\% & 75\% & 25\% & 50\% & 75\% \Tstrut\Bstrut\\[0.1cm]
    \hline
     1  & 13.73 & 12.08 &  9.68 & 11.51 & 10.14 &  8.39 \TTstrut\\[0.25cm]
     2  & 10.40 &  9.27 &  7.94 &  8.60 &  7.95 &  6.83         \\[0.25cm]
     3  & 17.59 & 16.92 & 15.66 & 16.32 & 15.45 & 13.96         \\[0.25cm]
     4  & 24.46 & 22.58 & 20.56 & 21.58 & 20.19 & 17.79         \\[0.25cm]
     5  & 18.52 & 16.88 & 13.99 & 14.63 & 12.99 & 10.96         \\[0.25cm]
     6  & 19.01 & 17.23 & 14.50 & 16.63 & 15.44 & 13.24         \\[0.25cm]
     7  & 25.73 & 23.14 & 19.30 & 19.81 & 18.62 & 15.96         \\[0.25cm]
     8  & 15.55 & 14.08 & 11.97 & 12.47 & 11.45 &  9.79         \\[0.25cm]
     9  & 19.34 & 17.73 & 15.41 & 16.81 & 15.43 & 13.17         \\[0.25cm]
    10  & 24.31 & 22.63 & 19.68 & 21.07 & 19.53 & 16.78         \\[0.25cm]

    \end{tabular}
\end{table}

Across all samples and sparsity levels, RTL consistently achieves higher PSNR than the single-model IMP baseline. The performance gap is observed for images with diverse scene content and appearance characteristics, confirming that the aggregate improvements reported in the main text are not driven by a small subset of favorable samples.

The advantage of RTL is particularly pronounced at higher sparsity levels, where enforcing a single global pruning mask across semantically heterogeneous regions leads to larger reconstruction errors. In contrast, RTL preserves region-specific parameters through specialized subnetworks, resulting in more stable degradation as sparsity increases.

\begin{figure}[h]
    \centering
    \includegraphics[width=1\columnwidth]{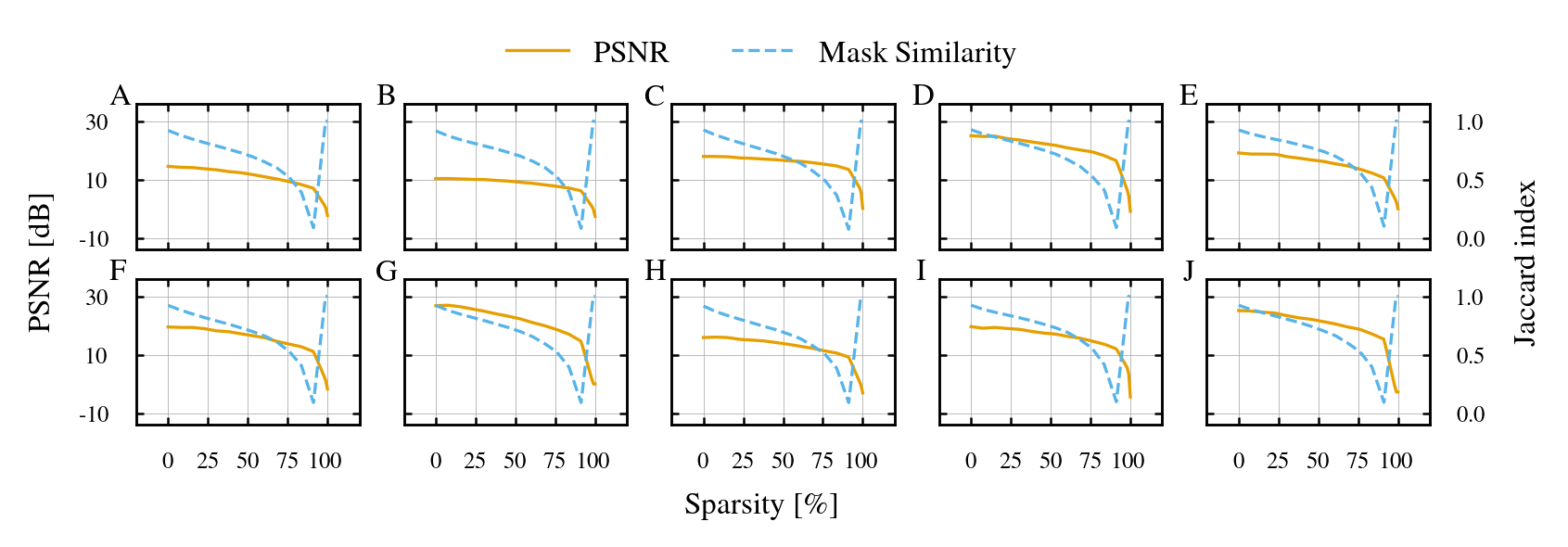}
    \caption{Per-image relationship between reconstruction quality and pruning mask similarity in the INR experiment. For each ADE20K image (A–J), PSNR (orange, left axis) and average mask similarity measured by the Jaccard index (blue dashed, right axis) are shown as a function of sparsity.}
    \label{fig:image_psnr}
\end{figure}

These per-image results complement the averaged PSNR values reported in the main text by demonstrating that the benefits of adaptive pruning in INR settings are consistent across images, rather than arising from outliers or dataset bias.
Figure~\ref{fig:image_psnr} shows the relationship between reconstruction quality and pruning mask similarity for each of the 10 ADE20K images (A–J) used in the INR experiment. For each image, we report PSNR (orange solid line, left axis) and average mask similarity measured by the Jaccard index (blue dashed line, right axis) as a function of sparsity.

Across all images, PSNR decreases gradually as sparsity increases, followed by a sharper drop at high sparsity levels. A similar trend is observed for mask similarity: pruning masks remain relatively stable at low to moderate sparsity, but their similarity rapidly decreases beyond a critical sparsity threshold. This transition is consistent across images, despite substantial differences in scene content and appearance.

Notably, the sharp decline in PSNR closely coincides with the point at which mask similarity collapses. This indicates that reconstruction quality degrades most significantly once subnetworks corresponding to different semantic regions begin to overlap or interfere, rather than as a direct consequence of parameter removal alone. At very high sparsity, mask similarity approaches zero, reflecting near-disjoint or unstable subnetworks and resulting in poor reconstruction quality.

These per-image trends mirror the averaged results reported in the main text and further support the interpretation that maintaining distinct, region-specific pruning masks is critical for preserving reconstruction performance in high-sparsity INR settings.

\subsection{INR subnetwork similarity vs. reconstruction quality}
\label{app:inr_similarity_psnr}

We further analyze the relationship between subnetwork specialization and reconstruction quality in the INR setting by examining how per-region PSNR correlates with mask similarity.

\begin{figure}[b!]
    \centering
    \includegraphics[width=1\columnwidth]{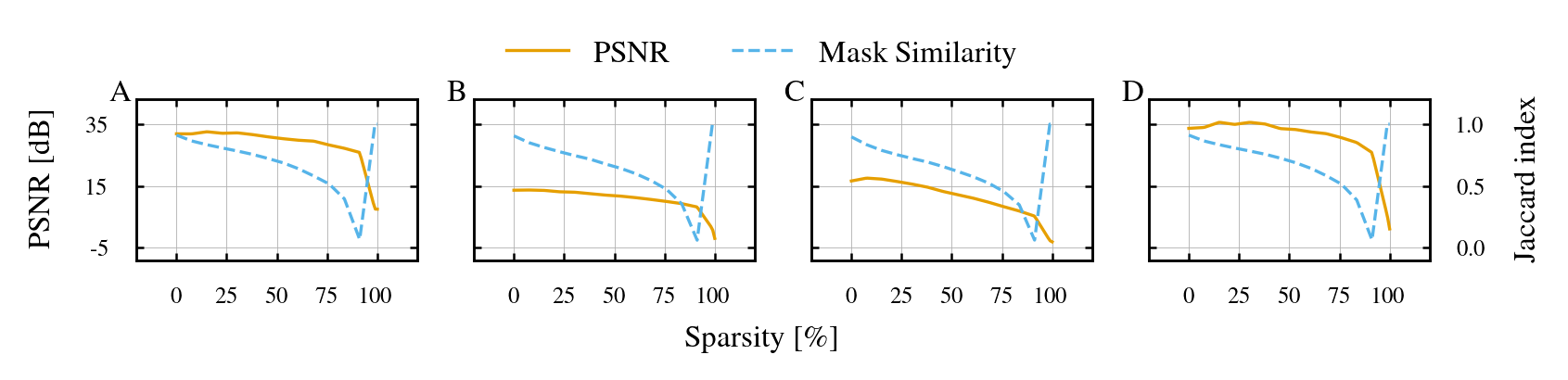}
    \caption{Per-class PSNR and subnetwork similarity for a single ADE20K image with four semantic regions. PSNR (solid) and average mask similarity to all other region-specific subnetworks (dashed) are shown as functions of sparsity. Performance degradation coincides with a rapid increase in mask similarity, indicating subnetwork collapse at high pruning ratios.}
    \label{fig:psnr_per_class_img0}
\end{figure}

Fig.~\ref{fig:psnr_per_class_img0} reports results for a single ADE20K image containing four semantic regions. For each region-specific subnetwork, we plot PSNR together with the average Jaccard similarity of its pruning mask to all other region masks across sparsity levels. PSNR remains relatively stable while mask similarity is low, but degrades sharply once similarity increases, indicating the onset of subnetwork collapse.

\begin{figure}[h]
    \centering
    \includegraphics[width=1\columnwidth]{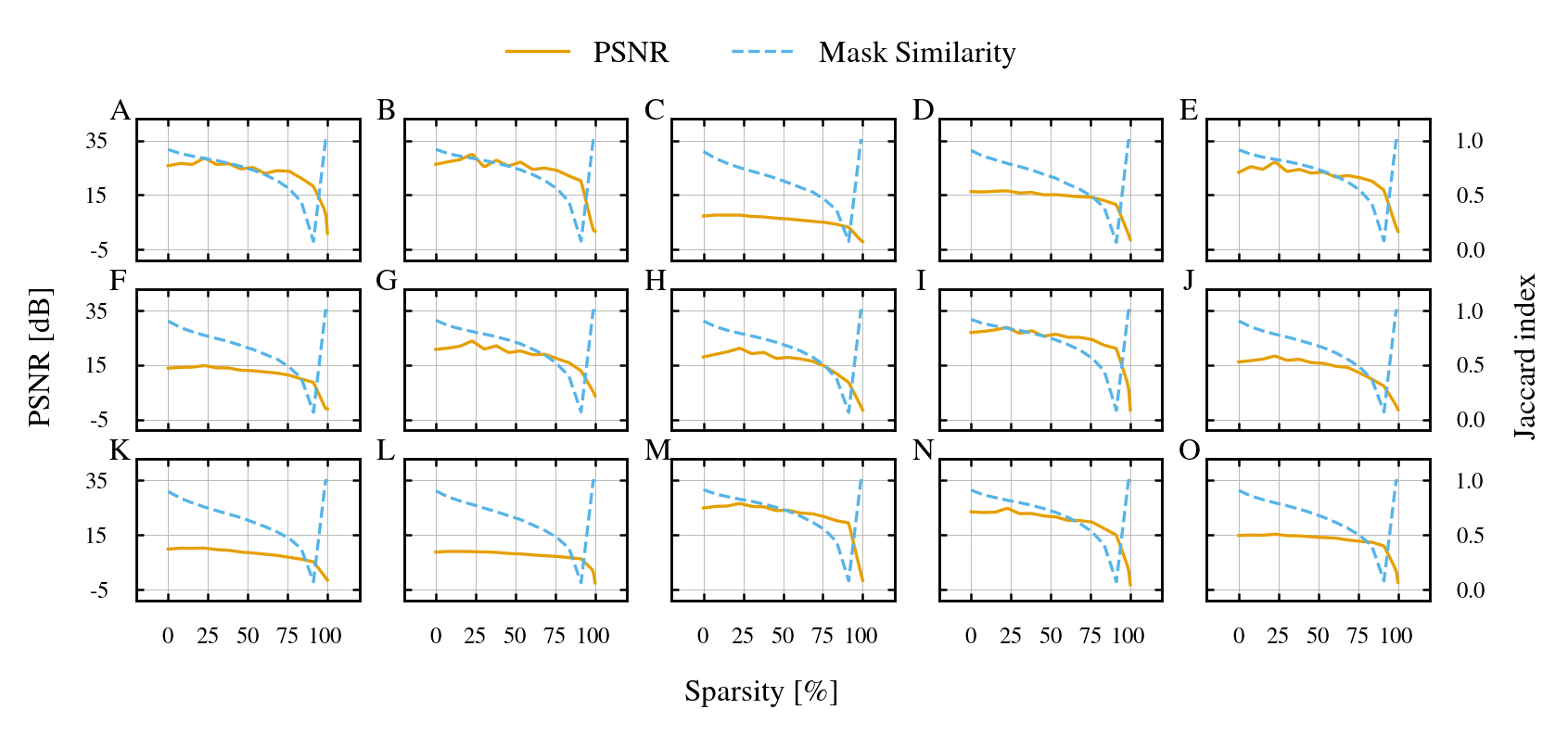}
    \caption{Per-class PSNR and subnetwork similarity for a second ADE20K image with fifteen semantic regions. Each subplot (A-O) corresponds to one semantic region. As sparsity increases, PSNR declines gradually until a sharp drop aligns with increasing similarity between region-specific pruning masks, reflecting loss of structural specialization.}
    \label{fig:psnr_per_class_img1}
\end{figure}

Fig.~\ref{fig:psnr_per_class_img1} extends this analysis to a second ADE20K image with fifteen semantic regions. Despite the larger number of classes, the same trend holds across all regions (A-O): reconstruction quality deteriorates rapidly once subnetworks lose structural distinctiveness under aggressive pruning.

\begin{figure}[b]
    \centering
    \includegraphics[width=1\columnwidth]{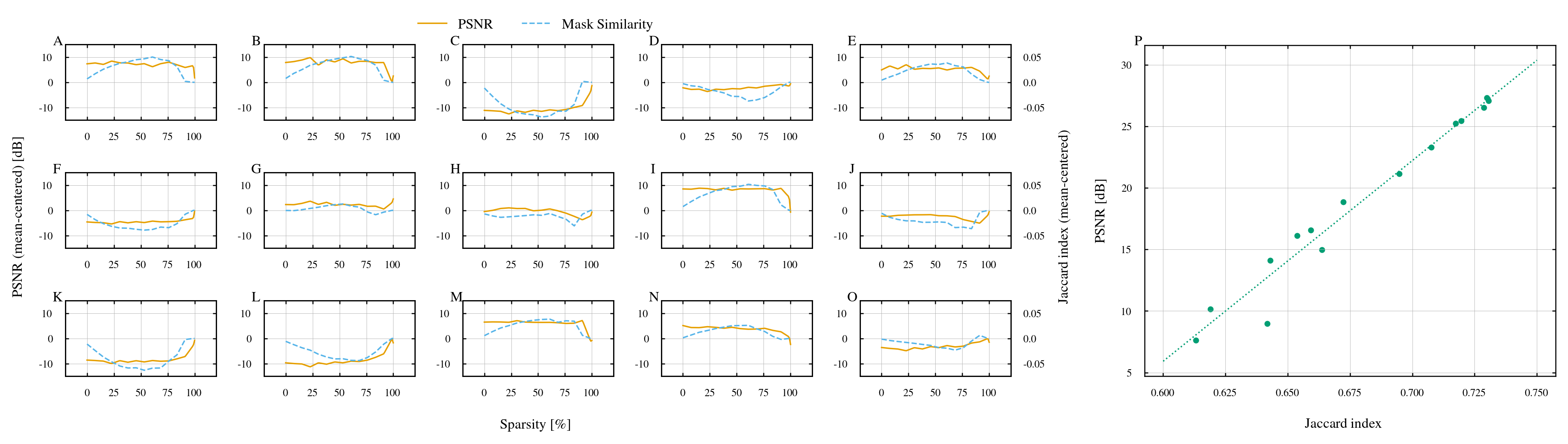}
    \caption{Correlation between reconstruction quality and subnetwork similarity in INRs.
Panels A–O show mean-centered PSNR and mean-centered mask similarity for the fifteen-region image. Panel P reports linear regression between the two quantities at 50\% sparsity across regions, revealing a strong monotonic relationship (Spearman $\rho \approx 0.982$).}
    \label{fig:psnr_per_class_img1_correlation}
\end{figure}

To isolate correlation effects independent of absolute scale, Fig.~\ref{fig:psnr_per_class_img1_correlation} re-plots the 15-region case with both PSNR and mask similarity mean-centered (panels A-O). Across regions, fluctuations in reconstruction quality closely track changes in mask similarity, revealing an inverse specialization effect: regions that maintain higher PSNR tend to exhibit greater mask overlap, while regions with lower PSNR show increased structural specialization. This suggests that well-reconstructed regions rely on more shared parameters, whereas harder regions benefit from stronger subnetwork differentiation.

Panel P summarizes this relationship at 50\% sparsity by plotting PSNR against mask similarity for all regions and fitting a linear regression. We observe a strong monotonic relationship, with an overall Spearman rank correlation of $\rho \approx 0.982$. This confirms that subnetwork similarity is a strong predictor of reconstruction quality in INRs, even within a single image.

Together, these results reinforce the central claim of RTL: preserving subnetwork distinctiveness is critical not only across datasets or tasks, but also for fine-grained, within-image semantic specialization. Mask similarity thus provides a reliable, label-free diagnostic for identifying oversparsification and impending performance collapse in coordinate-based representations.

\subsection{Qualitative INR Reconstructions under High Sparsity}

\begin{figure}[H]
    \centering
    \includegraphics[width=0.5\columnwidth]{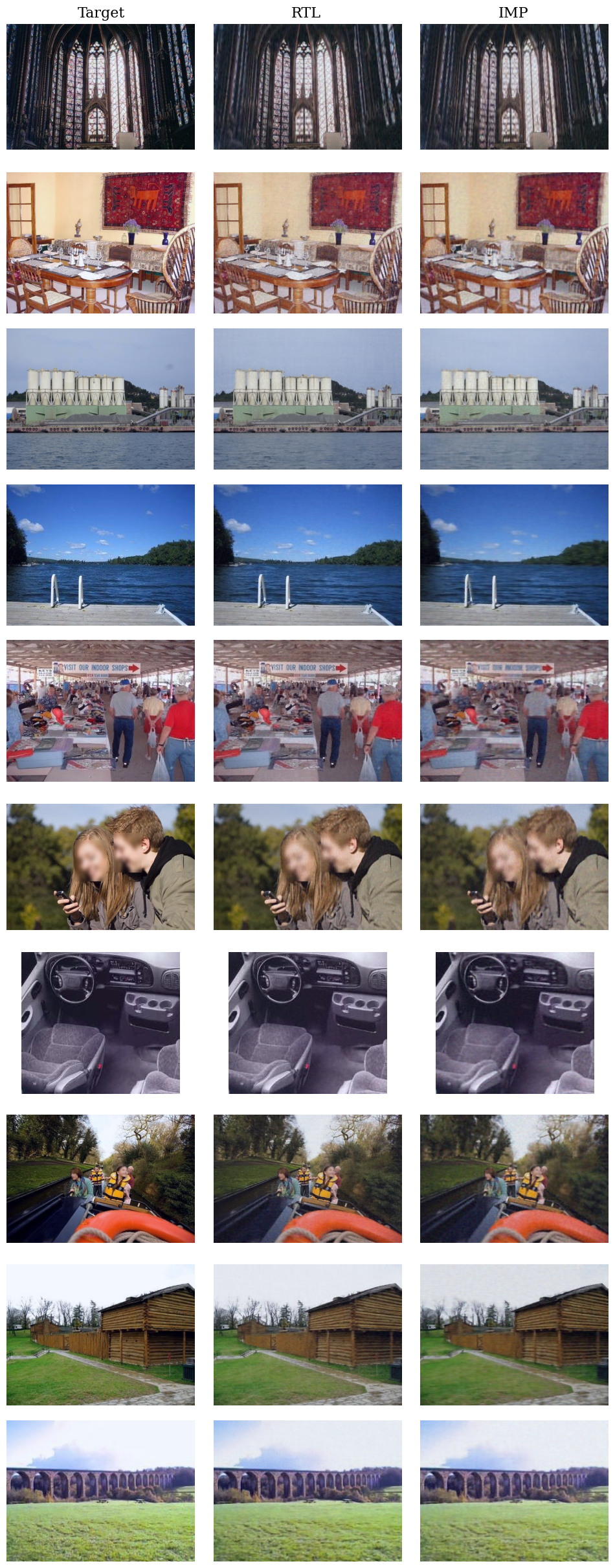}
    \caption{Qualitative INR reconstructions at 50\% sparsity. Rows correspond to 10 ADE20K images. Columns show the ground-truth target, RTL reconstruction, and IMP reconstruction. RTL preserves sharper edges and finer details compared to IMP, particularly in textured and semantically complex regions.}
    \label{fig:inr_vis_50}
\end{figure}

\begin{figure}[H]
    \centering
    \includegraphics[width=0.5\columnwidth]{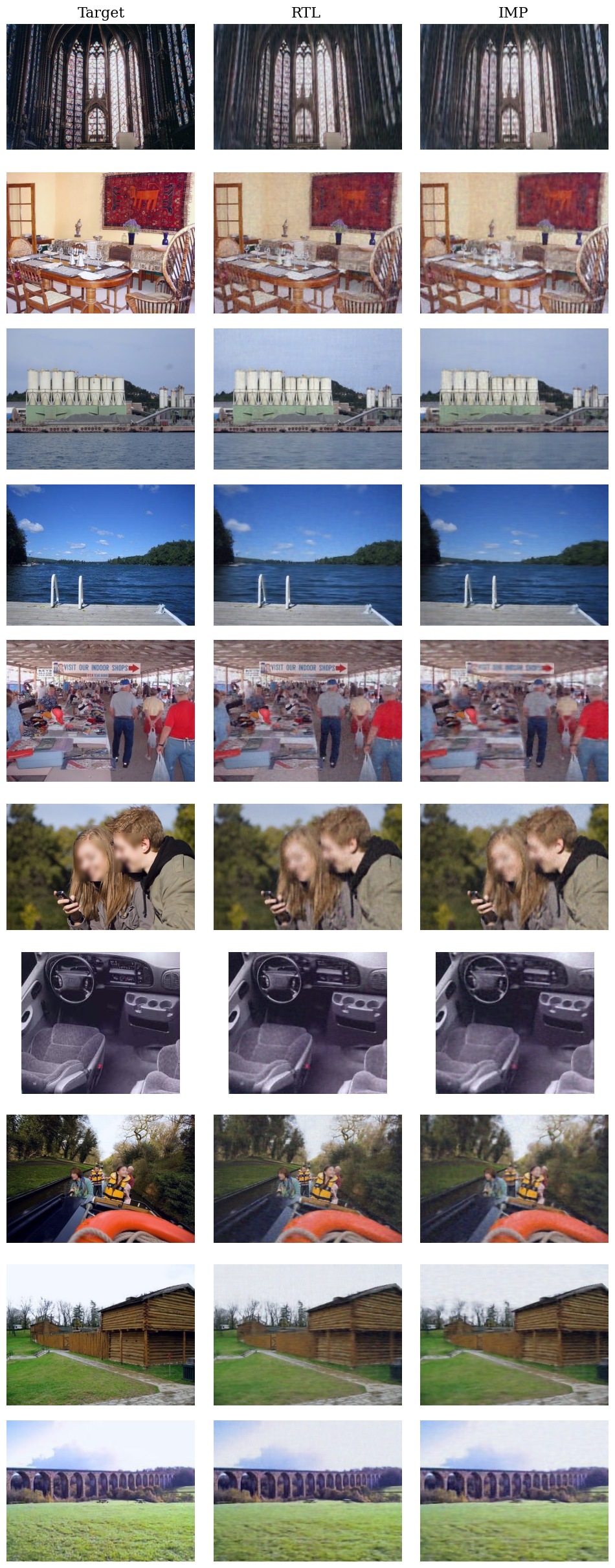}
    \caption{Qualitative INR reconstructions at 75\% sparsity. RTL maintains more faithful structure and color consistency, while IMP exhibits increased blurring and loss of high-frequency details. The gap between methods becomes more visible as sparsity increases.}
    \label{fig:inr_vis_75}
\end{figure}

\begin{figure}[H]
    \centering
    \includegraphics[width=0.5\columnwidth]{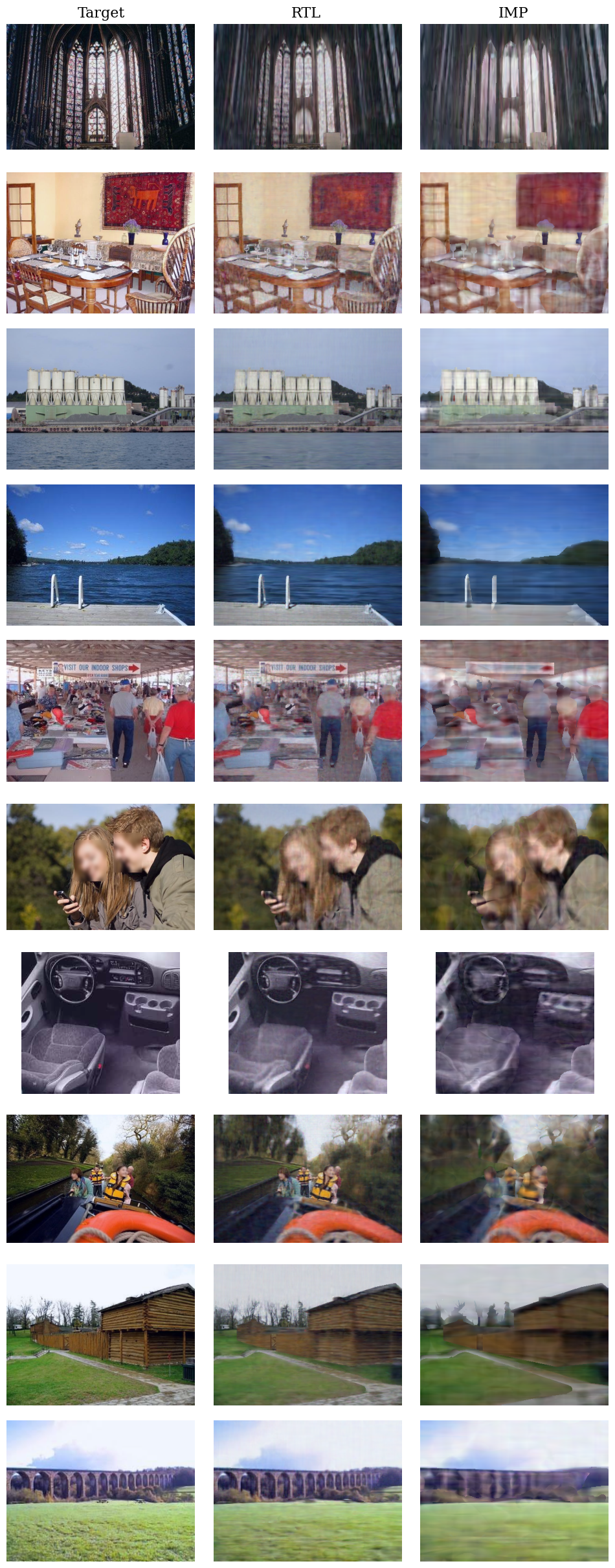}
    \caption{Qualitative INR reconstructions at 90\% sparsity. Under extreme pruning, IMP reconstructions often collapse to coarse approximations with severe artifacts. RTL degrades more gracefully, preserving recognizable object shapes and semantic boundaries across images.}
    \label{fig:inr_vis_90}
\end{figure}

To complement the quantitative PSNR analysis, we provide qualitative visualizations of INR reconstructions under increasing sparsity levels. Figures~\ref{fig:inr_vis_50}, \ref{fig:inr_vis_75}, and \ref{fig:inr_vis_90} show reconstructed images at 50\%, 75\%, and 90\% sparsity, respectively.

Each figure contains reconstructions for all 10 ADE20K images used in our experiments (rows). For each image, we display three columns: the ground-truth target image, the reconstruction produced by RTL, and the reconstruction produced by a single-mask IMP baseline. All models are trained under identical architectural, optimization, and sparsity constraints.

At 50\% sparsity (Fig.~\ref{fig:inr_vis_50}), both methods recover the overall scene structure, but RTL consistently preserves finer details and sharper object boundaries. Differences are particularly noticeable in textured regions and along semantic edges, where IMP reconstructions exhibit mild blurring and loss of contrast.

At 75\% sparsity (Fig.~\ref{fig:inr_vis_75}), the qualitative gap widens. IMP reconstructions show clear degradation in high-frequency content, with washed-out colors and incomplete reconstruction of small structures. In contrast, RTL maintains more faithful geometry and color consistency across most images, indicating that region-specialized subnetworks better preserve semantically important parameters under aggressive pruning.

At the extreme 90\% sparsity regime (Fig.~\ref{fig:inr_vis_90}), the difference becomes pronounced. IMP often collapses to coarse, low-detail approximations, with significant artifacts and loss of semantic coherence. RTL reconstructions, while degraded relative to lower sparsity levels, retain recognizable object shapes, clearer region boundaries, and more stable color distributions. This qualitative evidence aligns with the PSNR trends reported in the main paper, confirming that adaptive, region-specific pruning enables more graceful degradation as sparsity increases.

Overall, these visual results demonstrate that RTL not only improves average reconstruction metrics but also yields perceptually superior outputs, especially in the high-sparsity regime where competition for shared parameters is most severe.

\section{Speech Enhancement}
\label{app:speech_setup}

\subsection{Dataset construction}

We use clean speech samples from the DNS Challenge 2020 dataset \cite{reddy2020} and mix them with environmental noise from the TAU Urban Acoustic Scenes 2020 dataset \cite{heittola2020}. Noise samples are grouped into three acoustic scenes: \textit{indoor}, \textit{outdoor}, and \textit{transportation}. Each scene defines a distinct subset used to train a specialized subnetwork.

\subsection{Model architecture}

For the speech enhancement task, we use a lightweight U-Net–style architecture \cite{Ronneberger2015} operating on complex STFT representations. The network takes a 2-channel input (real and imaginary components) and predicts a 2-channel complex ratio mask.

The model consists of five encoder and five decoder stages. Each encoder stage applies a masked 2D convolution followed by ELU activation and batch normalization. Downsampling is performed only along the time axis using a stride of $(1,2)$, preserving frequency resolution. The encoder channel dimensions are $[12, 12, 12, 24, 48]$. The decoder mirrors the encoder using masked transposed convolutions with symmetric kernel sizes and strides. Skip connections concatenate encoder features with decoder inputs, doubling the channel dimensionality prior to decoding. The final decoder layer outputs a 2-channel complex mask.

All convolutions use $3\times3$ kernels, except for the first and last layers, which use $3\times5$ kernels to capture a wider temporal context. This compact design intentionally limits model capacity, allowing improvements from adaptive pruning to be isolated from architectural scaling effects. A detailed layer-by-layer specification of the network is provided in Table~\ref{tab:se_architecture} for reproducibility.

\begingroup
\setlength{\tabcolsep}{10pt}
\renewcommand{\arraystretch}{1.5}
\begin{table}[h]
\centering
\small
\caption{Speech enhancement U-Net architecture. All layers use ELU activation and batch normalization. Stride $(1,2)$ downsamples only along the time axis.}
\label{tab:se_architecture}
\begin{tabular}{c|c|c|c}
\hline
\textbf{Stage} & \textbf{Type} & \textbf{Channels (in $\rightarrow$ out)} & \textbf{Kernel / Stride} \\
\hline
Enc-1 & Conv2D & $2 \rightarrow 12$ & $3\times5$ / $(1,2)$ \\
Enc-2 & Conv2D & $12 \rightarrow 12$ & $3\times3$ / $(1,2)$ \\
Enc-3 & Conv2D & $12 \rightarrow 12$ & $3\times3$ / $(1,2)$ \\
Enc-4 & Conv2D & $12 \rightarrow 24$ & $3\times3$ / $(1,2)$ \\
Enc-5 & Conv2D & $24 \rightarrow 48$ & $3\times3$ / $(1,2)$ \\
\hline
Dec-1 & TConv2D & $48 \rightarrow 24$ & $3\times3$ / $(1,2)$ \\
Dec-2 & TConv2D & $48 \rightarrow 12$ & $3\times3$ / $(1,2)$ \\
Dec-3 & TConv2D & $24 \rightarrow 12$ & $3\times3$ / $(1,2)$ \\
Dec-4 & TConv2D & $24 \rightarrow 12$ & $3\times3$ / $(1,2)$ \\
Dec-5 & TConv2D & $24 \rightarrow 2$ & $3\times5$ / $(1,2)$ \\
\hline
\end{tabular}
\end{table}
\endgroup

\subsection{Signal processing and loss}

Input audio consists of 10-second, 16-bit waveforms. We compute STFTs using a 1024-sample window and a 256-sample hop size, producing spectrograms of shape $[2, 626, 513]$. The model predicts a mask of identical shape, which is applied to the noisy spectrogram.

Training uses the weighted source-to-distortion ratio (wSDR) loss. Unlike the vision experiments, this task does not involve negative samples; consequently, all methods are trained for the same number of epochs.

\subsection{Optimization and runtime}

We use the Adam optimizer with a learning rate of $1\mathrm{e}{-4}$ and no weight decay. Both pruning and retraining phases run for 10 epochs. All models share identical initial weights. Experiments are conducted on a single NVIDIA H100 GPU. Full RTL and multi-model IMP runs take approximately 10 hours, while the single-model IMP baseline completes in about 8 hours.

\subsection{Qualitative Analysis of Speech Enhancement}

\begin{figure}[t]
    \centering
    \includegraphics[width=0.9\columnwidth]{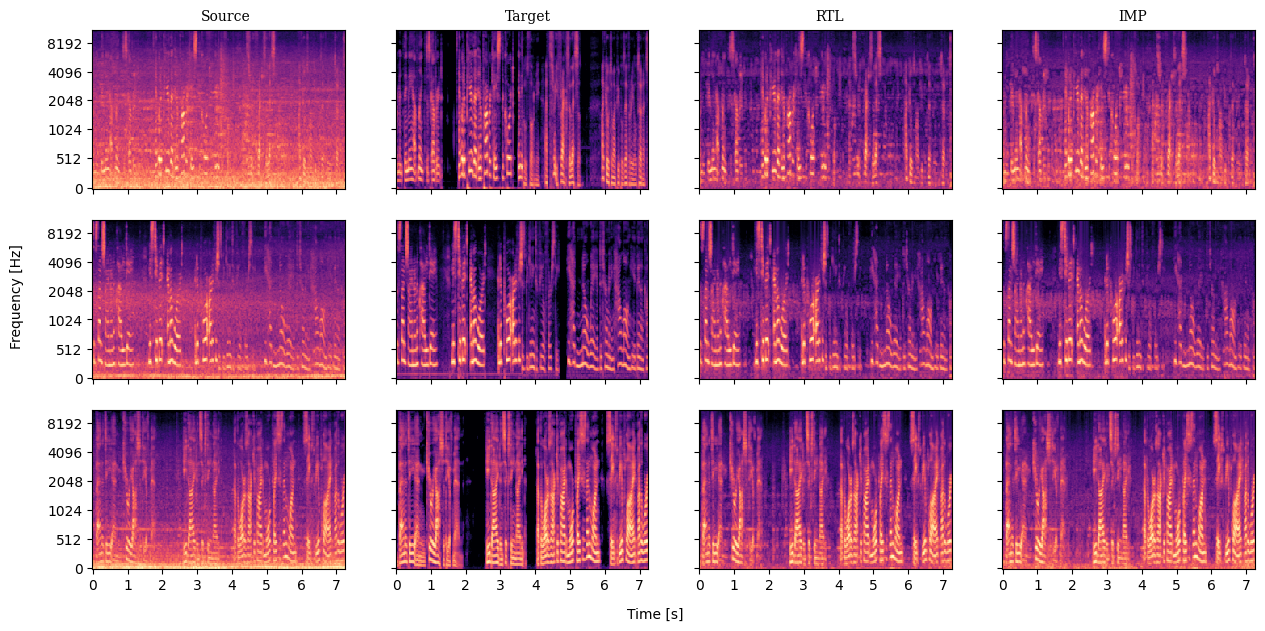}
    \caption{Qualitative speech enhancement results at 50\% sparsity. Each row corresponds to a different acoustic environment, and columns show the noisy input, clean target, RTL output, and IMP output. RTL more effectively suppresses noise and preserves harmonic speech structure across all environments.}
    \label{fig:se_spectrograms}
\end{figure}

Fig.~\ref{fig:se_spectrograms} provides a qualitative comparison of speech enhancement results across three acoustic environments at 50\% weight sparsity. Each row corresponds to a distinct environment class, while columns show the noisy input spectrogram, the clean target, and the enhanced 
outputs produced by RTL and the IMP baseline.

Across all environments, RTL more faithfully reconstructs the time-frequency structure of clean speech. Harmonic components are clearer, transient events are better preserved, and noise-dominated regions are more effectively suppressed. In contrast, the IMP baseline exhibits residual noise, smeared harmonics, and reduced contrast between speech and background, particularly in mid- and high-frequency bands.

The qualitative gap becomes especially apparent in challenging conditions, where environment-specific noise patterns dominate the input. RTL subnetworks, specialized to each acoustic scene, recover speech structure with higher temporal coherence and sharper spectral detail, whereas the single-mask IMP model struggles to balance denoising across heterogeneous conditions.

These visual results are consistent with the quantitative SI-SNRi improvements reported in Sec.~\ref{sec:results}, and further support the claim that adaptive, environment-aligned pruning yields more effective representations than a single global sparse model.

\section{Subnetwork Similarity Analysis}
\label{app:analysis}

\subsection{Mask similarity}

To analyze relationships between learned subnetworks, we compute pairwise similarity between binary pruning masks at multiple sparsity levels. Given two masks $M_i$ and $M_j$, similarity is measured using the Jaccard coefficient:
\begin{equation}
J(M_i, M_j) = \frac{|M_i \cap M_j|}{|M_i \cup M_j|},
\end{equation}
where intersection and union are computed element-wise over all prunable parameters.

Similarity is evaluated both globally and on a per-layer basis to study how overlap varies across network depth.

\subsection{Collapse analysis}

For each subnetwork, we record balanced accuracy at multiple sparsity levels and compute its average pairwise mask similarity to other subnetworks. Correlating these quantities allows us to assess how excessive overlap (i.e., subnetwork collapse) relates to performance degradation.

\subsection{Semantic alignment}

To assess whether structural similarity reflects semantic similarity, we use CIFAR-10 as a case study. Pairwise semantic distances between class labels are computed using WordNet path similarity \cite{miller1995wordnet}. These distances are compared against corresponding mask similarity values, yielding aligned similarity matrices that reveal whether conceptually related classes share pruning structure.


\end{document}